%% file: main.tex
\PassOptionsToPackage{svgnames,table}{xcolor}
\documentclass[pmlr,twocolumn,10pt]{jmlr}

\mlhtrack{proceedings}

\newif\iffinal
\finaltrue

\iffinal
    \ifmlhneedspmlr
      \jmlrvolume{346}
      \jmlryear{2026}
    \fi
    \ifmlhfindings \jmlrproceedings{}{ML4H 2026 - Findings Track}\fi
    \ifmlhdemo     \jmlrproceedings{}{ML4H 2026 - Demo Track}\fi
    \jmlrworkshop{Machine Learning for Health (ML4H) 2026}
\else
    \jmlrproceedings{}{Submitted to ML4H 2026: \mlhtrackname}
    \jmlrworkshop{Machine Learning for Health (ML4H) 2026}
\fi

\usepackage{array}
\usepackage{booktabs}

\newcommand{\equal}[1]{{\hypersetup{linkcolor=black}\thanks{#1}}}

\usepackage[utf8]{inputenc}
\usepackage[T1]{fontenc}
\usepackage{amsfonts}
\usepackage{graphicx}
\usepackage{xcolor}
\usepackage{natbib}
\usepackage{xurl}
\usepackage{listings}
\usepackage{csquotes}
\usepackage{microtype}
\usepackage{eurosym}

\newcommand{\rowrule}{\arrayrulecolor{black!25}\midrule[0.3pt]\arrayrulecolor{black}}

\definecolor{figblue}{HTML}{0072B2}
\definecolor{figgreen}{HTML}{009E73}

\definecolor{codebg}{RGB}{246,247,250}
\definecolor{codeframe}{RGB}{205,213,225}
\definecolor{codekw}{RGB}{120,60,140}
\definecolor{codestr}{RGB}{100,120,40}
\definecolor{codecmt}{RGB}{120,130,145}

\lstdefinestyle{codeblock}{
  basicstyle=\ttfamily\footnotesize,
  backgroundcolor=\color{codebg},
  frame=single,
  rulecolor=\color{codeframe},
  framerule=0.4pt,
  framesep=4pt,
  xleftmargin=8pt,
  xrightmargin=4pt,
  aboveskip=6pt,
  belowskip=6pt,
  breaklines=true,
  breakatwhitespace=true,
  columns=fullflexible,
  keepspaces=true,
  showstringspaces=false,
  captionpos=b,
}

\lstdefinestyle{pythoncode}{
  style=codeblock,
  language=Python,
  keywordstyle=\color{codekw}\bfseries,
  commentstyle=\color{codecmt}\itshape,
  stringstyle=\color{codestr},
  morekeywords={BaseModel,Field,Literal,ConfigDict,SystemName,ExamStatus,Timepoint,VitalSigns,SystemAssessment,PhysicalExamEvent,PhysicalExams,DischargeDiagnosis,Weight,OxygenSupport,OxygenSupportMethod,REDACTED},
}

\lstdefinestyle{promptblock}{
  style=codeblock,
  basicstyle=\ttfamily\footnotesize,
  literate={->}{{\textrightarrow}}1,
}

\title[MIMIC-Note-Extract]{MIMIC-Note-Extract: A Large-Scale Structured Dataset Extracted from MIMIC-IV-Note Discharge Summaries}

\author{%
  \Name{Kristian Schwethelm}\equal{These authors contributed equally} \Email{k.schwethelm@tum.de}\\
  \Name{Jonas Kuntzer}\footnotemark[1] \Email{jonas.kuntzer@tum.de}\\
  \addr Chair for AI in Healthcare and Medicine, TU Munich (TUM) and TUM University Hospital, Germany
  \AND
  \Name{Conrad Ketzer} \Email{conrad.ketzer@tum.de}\\
  \addr Department for Trauma Surgery, TUM University Hospital and TU Munich, Germany
  \AND
  \Name{Eva Wende} \Email{eva.wende@tum.de}\\
  \addr Department for Diagnostic and Interventional Radiology, TUM University Hospital and TU Munich, Germany
  \AND
  \Name{Moritz Schlager} \Email{moritz.schlager@tum.de}\\
  \addr Chair for AI in Healthcare and Medicine, TU Munich and TUM University Hospital, Germany\\
  \addr Munich Center for Machine Learning (MCML), Germany
  \AND
  \Name{Daniel Rückert} \Email{daniel.rueckert@tum.de}\\
  \addr Chair for AI in Healthcare and Medicine, TU Munich and TUM University Hospital, Germany\\
  \addr Department of Computing, Imperial College London, UK\\
  \addr Munich Center for Machine Learning (MCML), Germany
  \AND
  \Name{Georgios Kaissis} \Email{georg.kaissis@hpi.de}\\
  \addr Hasso Plattner Institute for Digital Engineering, University of Potsdam, Germany
  \AND
  \Name{Friederike Jungmann} \Email{friederike.jungmann@tum.de}\\
  \addr Chair for AI in Healthcare and Medicine, TU Munich and TUM University Hospital, Germany\\
  \addr Department for Diagnostic and Interventional Radiology, TUM University Hospital and TU Munich, Germany
  \AND
  \Name{Paul Hager} \Email{paul.hager@tum.de}\\
  \addr Chair for AI in Healthcare and Medicine, TU Munich and TUM University Hospital, Germany
}

\begin{document}
\maketitle

\begin{abstract}
MIMIC is one of the most widely used public hospital datasets for machine learning on electronic health records, yet clinically important information, such as past medical history, allergies, and diagnostic findings, remains confined to free-text discharge summaries.
Studies requiring this information must therefore build their own extraction pipelines, duplicating substantial effort across projects.
We release \textsc{MIMIC-Note-Extract} (MNE), a large-scale dataset that uses a hybrid extraction pipeline combining deterministic parsing with schema-guided large language model generation to transform 16 sections from 300{,}720 discharge summaries into structured tables.
MNE adds information that has no structured MIMIC counterpart and enriches related tables with broader coverage and greater clinical detail, including 10 times more patient-history entries and medication histories for more than twice as many admissions.
Validation by medical doctors confirms extraction accuracy above 92\% across sections.
Using MNE, we construct a clinical decision-making benchmark for LLM agents, demonstrating how reusable structured representations of clinical narratives reduce data curation effort.
\end{abstract}


\ifmlhneedsstatements
\paragraph*{Data and Code Availability}
MNE is available on PhysioNet~\citep{mne-physionet,physionet} under the same credentialed-access controls as its source dataset, MIMIC-IV-Note v2.2~\citep{johnson2023mimicivnote}. Code for reproduction is available at \url{https://github.com/kschwethelm/mimic-note-extract}.

\paragraph*{Institutional Review Board (IRB)}
No IRB approval required. Our work is a secondary analysis of deidentified MIMIC-IV-Note v2.2 data~\citep{johnson2023mimicivnote} and involves no interaction with individuals and no access to identifiable private information.
\fi

\section{Introduction}
\label{sec:intro}

\begin{figure*}[t]
	\centering
	\includegraphics[width=\linewidth]{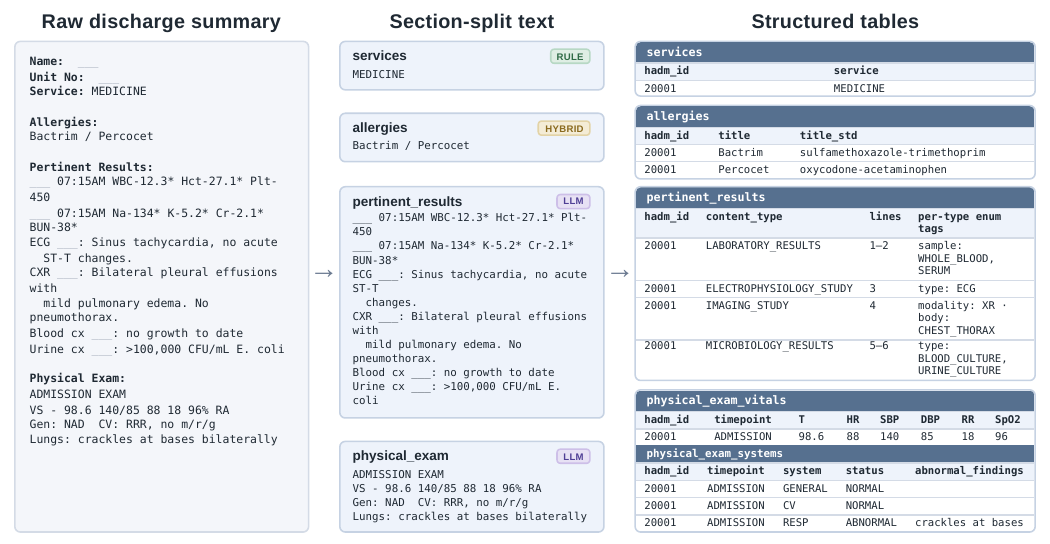}
	\caption{A raw discharge summary flowing through our extraction pipeline, showing a representative subset of four sections. The example is synthetic and does not correspond to any real admission. The discharge summary (left) is split into sections (middle) and decomposed into relational rows (right). Badges denote the extraction method of each section (rule-based, hybrid, or LLM).}
	\label{fig:intro}
\end{figure*}

The Medical Information Mart for Intensive Care (MIMIC)~\citep{johnson2023mimic} has enabled thousands of peer-reviewed publications~\citep{zhang2025datadriven}. However, the highly structured tabular formats that underpin much of this research omit the rich narrative context of patient care. Elements such as a patient's past medical history, allergies, and pertinent findings are only found in unstructured text. While MIMIC-IV-Note~\citep{johnson2023mimicivnote} provides these free-text narratives through a collection of clinical documents, including discharge summaries and radiology reports, this wealth of information remains difficult to scale or integrate into research workflows without extensive preprocessing.

This leads to a fragmented ecosystem of \enquote{individual preprocessing} where, in the absence of a standardized pipeline, each project must build its \emph{own} scripts to extract the specific variables needed for its use case. These isolated preprocessing pipelines are rarely shared or unified, resulting in duplicated engineering effort and inconsistent extraction quality across the community. But even if such an effort were made, the resulting datasets are typically tailored so closely to individual projects that they lack the generality needed to support broader research questions. At the core of this lies a methodological limitation: most existing projects focus on information \emph{extraction}. Instead, we posit that the community requires information \emph{conversion}---preserving the underlying information of the raw data while transforming it into an easily queryable, structured format. 

Our design philosophy reflects this distinction by prioritizing the completeness of the clinical record.
We introduce \textsc{MIMIC-Note-Extract} (MNE), a structured dataset derived from 300{,}720 MIMIC-IV-Note discharge summaries.
We employ a hybrid processing framework that pairs deterministic parsing with schema-guided LLM generation to translate 16 major clinical sections of the discharge summary into structured tables.
By providing structured representations for 13 of these sections---ranging from pre-admission history to detailed physical exams---we eliminate the preprocessing barrier while preserving the semantic nuances of the original narratives (see Figure~\ref{fig:intro}).

To demonstrate the dataset's utility, we construct a large-scale clinical decision-making benchmark that evaluates LLM agents on a real-world diagnostic task. It supports more pathologies and more specific clinical actions with substantially less curation effort.

\paragraph{Contributions.}
\begin{itemize}
	\item We introduce and publicly release \textsc{MIMIC-Note-Extract}, a clinical database containing data from 16 discharge summary sections of 300{,}720 admissions, validated via expert medical review to ${>}92\%$ accuracy and shown to complement existing MIMIC tables with novel information.
	\item Together with medical experts, we design an adaptive extraction pipeline that balances computational efficiency and clinical accuracy by dynamically choosing rule- or LLM-based extraction.
    \item We construct an agentic benchmark with 10{,}000 patient cases to evaluate LLMs' capabilities for diagnosing 19 different pathologies.
\end{itemize}

\input{sections/related_work}
\input{sections/dataset_pipeline}
\input{sections/validation}
\input{sections/comparison}
\input{sections/case_study}
\input{sections/discussion}
\input{sections/conclusion}

\acks{
KS and GK received support from the Bavarian Collaborative Research Project PRIPREKI of the Free State of Bavaria Funding Programme “Artificial Intelligence – Data Science”.

JK was supported by the German Medical Informatics Initiative grant ID 01ZZ22316C.

The authors gratefully acknowledge the scientific support and resources of the AI service infrastructure LRZ AI Systems provided by the Leibniz Supercomputing Centre (LRZ) of the Bavarian Academy of Sciences and Humanities (BAdW), funded by the Bayerisches Staatsministerium für Wissenschaft und Kunst (StMWK), primarily through the BayernKI initiative.
}

\bibliography{references}

\appendix
\onecolumn

\input{sections/appendix}

\end{document}

%% file: sections/related_work.tex
\section{Related Work}
\label{sec:related}

\paragraph{MIMIC-derived tabular datasets.}
Several works derive task-specific datasets by selecting, merging, and engineering features from MIMIC's existing structured tables: MIMIC-Extract builds time-series features from MIMIC-III chart events \citep{wang2020mimicextract}, the MIMIC-III multitask benchmark packages vital signs and lab results for clinical prediction \citep{harutyunyan2019multitask}, the MIMIC-IV emergency-department (ED) benchmark assembles triage outcomes from ED tables \citep{xie2022mimicived}, and \citet{gupta2022mimicivpipeline} standardize cohort construction across these sources.
Instead of selecting from existing tables, we convert \emph{free-text discharge summaries} into new structured tables, adding clinically important information that is absent from MIMIC's tabular data.

\paragraph{Radiology reports.}
Beyond the structured tables, MIMIC provides two free-text document types: radiology reports and discharge summaries \citep{johnson2023mimicivnote}.
The radiology reports are widely studied and have been structured into entity-relation graphs, standardized report formats, disease labels, or question-answer pairs \citep{jain2021radgraph,delbrouck2025srr,muller2026cxrqba}. We instead work on discharge summaries, which cover the whole admission rather than a single imaging study. However, our extraction can easily be combined with extractions of the radiology report sections.

\paragraph{Manual discharge-summary annotation.}
Several datasets share our focus on discharge summaries, but manually label passages in them for a single, narrow purpose: DiReCT \citep{wangMIMICIVExtDiReCT2025} marks findings that support a diagnosis to annotate the diagnostic reasoning chain, CLIP \citep{mullenbach2021clip} flags post-discharge to-dos (e.g., \enquote{follow up with cardiology in two weeks}), MedDec \citep{elgaar2024meddec} marks where a clinical decision was made (e.g., \enquote{started IV antibiotics}), and the 2018 n2c2 task \citep{n2c2} tags medications and the adverse drug events they cause.
Because manual annotation is costly, each covers only a few hundred notes (718 for CLIP, 511 for DiReCT, and 451 for MedDec), orders of magnitude smaller than our cohort.

\paragraph{Automated discharge-summary extraction.}
Prior automated approaches extract text from the notes without structuring it: MIMIC-IV-Ext-BHC \citep{mimic_iv_ext_bhc} releases the verbatim brief-hospital-course section of 270{,}033 summaries as a summarization target, clinical decision-making (CDM) benchmarks \citep{hagerEvaluationMitigationLimitations2024,gaber2025mimicivcds,ferber2026mira} use regular expressions (regex) to assemble diagnostic cases from selected note sections such as the patient history (500--9k cases), and MIMIC-IV-Ext-22MCTS \citep{wang2025mimic22mcts} uses Llama-3.1-8B to decompose ${\sim}267$k summaries into a list of short clinical events (e.g., \enquote{fever} or \enquote{pain control}).
In contrast, we do not target a single task, section, output type, or extract only verbatim text: we convert \emph{all 16} sections of every discharge summary across ${\sim}300$k admissions into general, structured tables with well-defined fields, creating a general-purpose dataset that supports a wide range of downstream applications.

%% file: sections/dataset_pipeline.tex
\section{Dataset and Extraction Pipeline}
\label{sec:dataset}

\begin{table*}[tb]
	\centering
	\footnotesize
	\caption{Clinical sections, key concepts, extraction methodologies, and cohort-wide coverage in MNE across 300{,}720 admissions. Coverage is the percentage of admissions with a non-empty section.}
	\label{tab:schema_and_coverage}
	\begin{tabular*}{\textwidth}{@{\extracolsep{\fill}}lp{5.5cm}lrr}
		\toprule
		Clinical Section            & Key Fields                                                                                       & Method   & Admissions & Coverage \\
		\midrule
		Gender                      & Patient sex assigned at birth                                                                               & Rule     & 300{,}720  & 100.0\%  \\
		Services                    & Clinical service at discharge                                                                              & Rule     & 300{,}720  & 100.0\%  \\
		Chief Complaint             & Symptom category and clinical description                                                                   & Hybrid   & 296{,}132  & 98.5\%   \\
		History of Present Illness  & Chronological narrative of events prior to admission                                                        & Verbatim & 296{,}620  & 98.6\%   \\
		Past Medical History        & Prior conditions and interventions by category, chronicity, and subject                                     & LLM      & 290{,}212  & 96.5\%   \\
		Medications on Admission    & Active home medications with dose, route, frequency, and strength                                           & LLM      & 275{,}035  & 91.5\%   \\
		Allergies                   & Allergen name                                                                                               & Hybrid   & 296{,}378  & 98.6\%   \\
		Physical Examination       & Physiological measurements and findings by organ system                                                     & LLM      & 294{,}708  & 98.0\%   \\
		Pertinent Results           & Lab, imaging, and diagnostic findings with key values                                                       & LLM      & 241{,}508  & 80.3\%   \\
		Brief Hospital Course       & Narrative of clinical course and medical management                                                         & Verbatim & 267{,}164  & 88.8\%   \\
		Major Procedures            & Procedural interventions with current vs. historical flags                                                  & Hybrid   & 159{,}791  & 53.1\%   \\
		Discharge Diagnosis         & Discharge diagnoses preserving clinical priority                                                            & Hybrid   & 297{,}537  & 98.9\%   \\
		Discharge Condition         & Clinical status indicators                                                                                  & Hybrid   & 300{,}109  & 99.8\%   \\
		Discharge Medications       & Active outpatient prescriptions at discharge with dose, route, and frequency                                & LLM      & 285{,}926  & 95.1\%   \\
		Discharge Instructions      & Outbound instructions, follow-up directives, and education                                                  & Verbatim & 291{,}647  & 97.0\%   \\
		Discharge Disposition       & Discharge destination                                                                                       & Rule     & 300{,}667  & 100.0\%  \\
		\bottomrule
	\end{tabular*}
\end{table*}

We derive our cohort from the 331{,}793 raw discharge summaries available in MIMIC-IV-Note v2.2, which are released de-identified, with sensitive information redacted by \enquote{\_\_\_}.
The redaction method also removed some non-identifying clinical content, including clinical terms and parts of section headers.
The remaining text retains all patient histories, symptoms, diagnoses, treatments, findings, and descriptions of the hospital course, although some keywords are lost through redaction.
Following quality filtering\footnotemark, our final dataset comprises 300{,}720 admissions.
\footnotetext{Our filtering pipeline removes non-discharge summary note types, formatting outliers (such as specific external dictation system templates), and reports with disordered or severely incomplete clinical sections.}
The resulting \textsc{MIMIC-Note-Extract} dataset preserves high data availability across nearly all clinical categories (Table~\ref{tab:schema_and_coverage}), with most sections exceeding 90\% coverage. 
Coverage denotes the percentage of admissions with a non-empty section; for example, the low coverage of \enquote{Major Procedures} directly reflects the availability of procedure information in the source data.
We also preserve three high-complexity narrative sections in their original form to avoid losing information where a structured breakdown was not feasible.
The following paragraphs detail our extraction pipeline; the full database schema and comprehensive technical details are provided in Appendices~\ref{app:full-schema} and~\ref{app:pipeline-details}.

\paragraph{Splitting into sections.}
The first step in our pipeline divides monolithic discharge reports into their constituent sections.
We identify 24 distinct sections, dropping 8 that are fully redacted (e.g., \enquote{Name} in Fig. \ref{fig:intro}).
De-identification is, at times, over-inclusive and also removes tokens that carry no identifying information, including words within section headers.
Splitting is therefore complicated by partially redacted headers (e.g., \enquote{Discharge \_\_\_}, which could indicate \enquote{Discharge Condition} or \enquote{Discharge Disposition}).
To resolve this, our deterministic splitting algorithm tracks common header variants.
It further exploits the consistent relative ordering of clinical sections to infer headers that remain ambiguous due to redaction.\footnote{A remaining 5{,}461 ambiguous cases that could not be parsed automatically were manually resolved.}

\paragraph{Information modeling.}
Following section splitting, the next step in our pipeline is defining the target schemas for information extraction. The difficulty lies in the massive structural and content diversity within individual clinical sections. For instance, the \emph{Pertinent Results} section can range from a single-line negative cell culture report to a multi-page unfiltered output of microbiology panels and longitudinal imaging findings. Therefore, we must carefully decide what information to retain and how to structure it; the goal is to capture essential clinical depth without introducing hyper-specific fields that remain empty across most samples. Keeping the schemas concise also ensures they remain manageable for downstream human verification. To achieve this balance, we iteratively developed target models in close consultation with clinicians, analyzing source texts for recurring patterns to refine our field definitions.
The schemas preserve temporal context where it is available in the source, including distinctions between current and historical information.

\paragraph{Adaptive extraction.}
To process large volumes of clinical notes efficiently, we follow the principle of choosing the simplest tool capable of handling the task.
Simple, highly structured sections can be reliably processed using deterministic rules, such as regular expressions and string matching, whereas complex narratives require the contextual understanding of LLMs\footnotemark.
Our pipeline thus categorizes sections into three distinct operational modes (see Table \ref{tab:schema_and_coverage}): some rely entirely on \textit{rule-based} extraction, others require \textit{LLMs} exclusively, and a \textit{hybrid} group uses rule-based methods by default, falling back to an LLM only when section-specific validation criteria fail.
For example, for \emph{Discharge Condition}, we extract and validate status indicators deterministically; any residual text triggers LLM fallback (Appendix~\ref{app:pipeline-details}).
This adaptive approach ensures optimal computational throughput for straightforward data while reserving LLM inference for genuinely complex data.
Furthermore, we optimized token efficiency by mapping long clinical identifiers to single-token codes (e.g., mapping \texttt{subject} to \texttt{s}).
By instructing the model to output these short abbreviations, we trade off expensive output tokens for much cheaper input tokens.
This approach reduced overall inference costs (see Appendix~\ref{app:cost-optimization}).
Across the entire cohort, this pipeline was executed for a total cost of 1{,}894.61\euro, yielding substantial cost savings of 321.24\euro{} (14.5\% of total cost) compared to a naive LLM-only baseline (see Table~\ref{tab:unified_costs_appendix} in Appendix~\ref{app:cost-calculations}).

\footnotetext{We used Gemini 2.5 Flash~\citep{gemini-flash-25} and Gemini 3.1 Flash-Lite~\citep{gemini-flash-lite} through Google Cloud Vertex AI API with zero data retention; Appendix~\ref{app:cost-calculations} lists the model used for each section.}

\paragraph{Validation and constraints.}
To ensure the final output aligns with our target data models, the pipeline concludes with a strict validation layer. We use formal schemas to programmatically constrain LLM generation \citep{beurer2024guidingLLMs,banerjee2025crane} and to run post-hoc data validation. Any output that violates these structural constraints or custom validation rules triggers an automated retry loop, ensuring a guaranteed fit with the target data model.

%% file: sections/validation.tex
\section{Quality Validation}
\label{sec:validation}

We evaluate extraction quality through a manual, side-by-side review of the source text against the extractions by three medical doctors. Annotations were performed at the item level to allow the distinction between a single item-level error and an entirely incorrect sample.
The error taxonomy, concrete examples, and additional metrics are shown in Appendix~\ref{app:validation-full}.

\begin{table}[tb]
  \centering
  \small
  \caption{Extraction accuracy (item-level, 95\% confidence margins of error) and inter-annotator agreement (Gwet's AC1) per section.}
  \label{tab:validation}
  \input{tables/validation_summary.tex}
\end{table}

\paragraph{Setup.}
Rule-based extractions were excluded from manual review due to their deterministic, auditable nature. 
Final validation was performed after the extraction pipeline had been frozen.
Extractions reviewed during this validation were not manually corrected.
For LLM-based extractions, we determined the required sample size using the standard normal approximation \citep{lohr2021sampling} formula for binomial proportions. 
Assuming an expected accuracy of $p = 0.80$ at a 95\% confidence level and a conservative target margin of error (MoE) of 6\%, the formula dictated a minimum of 171 unique cases per section. 
To assess inter-annotator agreement (IAA), we utilized 121 single-annotated and 50 dual-annotated cases. 
The dual-annotated cases were randomly distributed among pairs of our three practitioners. 
This setup ensured that each overlapping case was independently reviewed by two people, resulting in a total workload of $\approx74$ cases per annotator  per section. 
In total, the annotators completed 2{,}210 case-level tasks (13{,}744 individual item-level decisions).
Alongside accuracy, we report Gwet's AC1~\citep{gwet2008ac1} for IAA, since raw agreement is inflated by the high prevalence of correct extractions.

\paragraph{Results.}
As summarized in Table~\ref{tab:validation}, item-level accuracy is consistently high across all sections, ranging from 92.2\% to 99.8\%, with realized margins of error between 0.32\% and 3.81\%, well inside our 6\% target.
IAA is similarly high: AC1 exceeds 95\% in seven of the ten sections.
The three exceptions are pertinent results at 88.4\%, past medical history at 90.1\%, and physical examination at 90.4\%.

%% file: tables/validation_summary.tex
  \begin{tabular}{lrr}
    \toprule
    \textbf{Section} & \textbf{\shortstack{Acc. (\%)}} & \textbf{\shortstack{AC1}} \\
    \midrule
    Allergies                  & 99.8 $\pm$ 0.32 & 99.8 \\
    Chief Complaint            & 97.6 $\pm$ 2.37 & 98.0 \\
    Discharge Condition        & 97.4 $\pm$ 3.15 & 98.0 \\
    Discharge Diagnosis        & 97.1 $\pm$ 2.65 & 95.8 \\
    Major Procedures           & 98.3 $\pm$ 2.04 & 98.0 \\
    Past Medical History       & 95.2 $\pm$ 1.48 & 90.1 \\
    Physical Examination       & 95.2 $\pm$ 1.83 & 90.4 \\
    Pertinent Results          & 92.2 $\pm$ 3.81 & 88.4 \\
    Discharge Medications      & 97.2 $\pm$ 1.17 & 97.1 \\
    Medications on Admission   & 99.3 $\pm$ 0.47 & 97.2 \\
    \bottomrule
  \end{tabular}

%% file: sections/comparison.tex
\section{Comparison to MIMIC Tables}
\label{sec:comparison}

\begin{table}[tb]
  \centering
  \footnotesize
  \caption{Relationship of extracted sections to MIMIC's existing tabular data.}
  \label{tab:mimic-comparison-overview}
  \setlength{\tabcolsep}{3pt}%
  \begin{tabular}{>{\raggedright\arraybackslash}p{0.27\linewidth}p{0.675\linewidth}}
    \toprule
    \textbf{Section} & \textbf{Contribution} \\
    \midrule
    \multicolumn{2}{l}{\emph{Already exists in MIMIC, but can flag mismatches}} \\
    \midrule
    Gender & Flag mismatches\\
    \rowrule
    Services & Flag mismatches \\
    \rowrule
    Discharge Disposition & Flag mismatches; adds ${\sim}44$k missing values \\
    \midrule
    \multicolumn{2}{l}{\emph{Related to MIMIC information, but broader or richer}} \\
    \midrule
    Discharge Diagnosis & Clinician-written diagnoses with primary/secondary structure \\
    \rowrule
    Chief Complaint & 2$\times$ more admissions; cleaner terms and additional clinical context \\
    \rowrule
    Allergies & Comprehensive allergen list covering far more admissions than MIMIC's ICD codes \\
    \rowrule
    Past Medical History & 10$\times$ more history rows; contains broader history, family, substance, and functional history, compared to MIMIC's ICD codes \\
    \rowrule
    Medications on Admission & $>2\times$ more admissions; more detail by providing dose, route, frequency, PRN, and hold status \\
    \rowrule
    Major Procedures & Clinical procedure label and more surgical details compared to ICD/HCPCS billing \\
    \midrule
    \multicolumn{2}{l}{\emph{No structured MIMIC counterpart}} \\
    \midrule
    Discharge Condition & Provides structured discharge state, including mental status and functional activity \\
    \rowrule
    Discharge Medications & Provides discharge-medication list \\
    \rowrule
    Physical Examination & Provides system-level examination findings with normal/abnormal status \\
    \rowrule
    Pertinent Results & Provides important clinical results, e.g., radiology interpretations \\
    \bottomrule
  \end{tabular}
\end{table}

The information we extract from the discharge notes overlaps partially with MIMIC's existing tables.
We compare each MNE section with its closest structured counterpart(s) in MIMIC to determine whether it validates, extends, or adds to existing data.
Exact one-to-one matching is unrealistic for most sections: MNE fields are extracted from unstructured text, whereas MIMIC tables contain coded events.

Table~\ref{tab:mimic-comparison-overview} groups the sections by how they relate to MIMIC's tables and gives a short summary of what each section contributes.
Three sections largely reproduce an existing MIMIC field; the redundancy can be used to flag mismatching records.
The discharge-disposition extraction additionally covers ${\sim}44$k admissions with a missing MIMIC value.
The remaining sections either extend a related but non-equivalent MIMIC table with broader coverage or richer detail (e.g., allergies, past medical history, medications on admission) or have no MIMIC counterpart at all (discharge condition, discharge medications, physical exam, and pertinent results).
Overall, MNE complements MIMIC's existing structured data by adding substantial new clinical information and providing broader or richer representations where related tables already exist.
See Appendix~\ref{app:mimic-comparison-details} for detailed results.

%% file: sections/case_study.tex
\section{Benchmarking Agents in Clinical Decision-Making}
\label{sec:case-study}

As a concrete downstream application of our dataset, we build an agentic benchmark for clinical decision-making (CDM) based on MIMIC-CDM~\citep{hagerEvaluationMitigationLimitations2024}. 
Such benchmarks are a natural application for our dataset because they require clinical information that is not available in MIMIC's tabular data.
MIMIC-CDM simulates real-world clinical decision-making from patient presentation to diagnosis: starting from patient history, the doctor agent decides which examinations and tests to perform before making a diagnosis.
Like a human doctor would examine a patient or order a laboratory test or imaging study, the agent initiates these actions through tools and receives the corresponding findings.
To supply these findings, the authors parsed discharge summaries with regex, a laborious process that yielded verbatim free-text blocks (similar to the \enquote{Section-split text} in Figure \ref{fig:intro}).
We ask how the construction of MIMIC-CDM would change if MNE had been available at the time.

We first validate that MNE can replace the regex-derived texts without reducing accuracy, while requiring far less curation effort. We then leverage MNE's fields to build a larger agentic CDM benchmark with more admissions and pathologies, stricter case filtering, and finer control over the agent's context.

\begin{table}[tbh]
  \centering
  \footnotesize
  \setlength{\tabcolsep}{3pt}
  \caption{Diagnosis accuracy (\%) on the full-information and agentic CDM benchmarks.
  Parentheses report how accuracy changes when using our data instead of the original regex-derived data.
  Appendix Table~\ref{tab:cdm-confidence-intervals} reports 95\% confidence intervals.}
  \label{tab:case-study-results}
  \begin{tabular}{@{}l l c c@{}}
    \toprule
    \textbf{Model} & \textbf{Reasoning} & \textbf{Full info} & \textbf{Agentic} \\
    \midrule
    Gemini 3.5 Flash & minimal & 88.7 (+0.1) & \textbf{87.9} \\
    Gemini 3.1 Flash Lite & minimal & 87.4 (+0.7) & 81.9 \\
    Gemini 3.1 Flash Lite & high & 88.6 (+0.0) & 84.2 \\
    Gemma 4 31B & $\times$ & 86.7 (+0.3) & 83.7 \\
    Qwen3.6 35B-A3B & $\times$ & \textbf{90.4} (+0.3) & 83.6 \\
    Qwen3.6 35B-A3B & $\checkmark$ & 89.3 (+0.2) & 82.6 \\
    Qwen3.5 2B & $\times$ & 82.0 (+0.9) & 51.3 \\
    \bottomrule
  \end{tabular}
\end{table}

\begin{figure*}[tb]
  \centering
  \includegraphics[width=.95\textwidth]{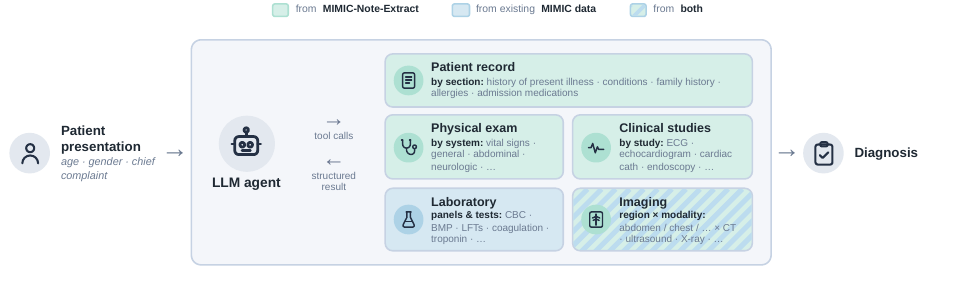}
  \caption{Overview of the agentic CDM benchmark built from \textsc{MIMIC-Note-Extract}.
  The LLM agent starts from the patient presentation and gathers evidence through five tools before making a diagnosis.
  \textcolor{figgreen}{Green} tools are sourced from MNE, \textcolor{figblue}{blue} tools from existing MIMIC data, and hatched from both.
  In contrast to prior benchmarks, the agent must name the exact item it wants, e.g., a single examination system or a patient-record section, and receives only this requested part.}
  \label{fig:benchmark}
\end{figure*}

\subsection{Full-Information CDM Benchmark}
\label{sec:case-study:reproduction}

We reproduce the full-information setting of MIMIC-CDM, in which the LLM receives all patient information up front instead of requesting it through tools.
This resembles a second-reader setting: the model interprets evidence already selected and collected by the treating clinicians to reach a diagnosis.
Specifically, the LLM receives a single message containing the patient's history of present illness, physical examinations, laboratory tests, and radiology reports.
This setting evaluates whether the model can reach the correct diagnosis when the relevant clinical information has already been selected.

As described above, \citet{hagerEvaluationMitigationLimitations2024} had to preprocess discharge summaries to construct this context.
We instead can query the relevant fields from MNE and apply simple filters where necessary.
For example, the \emph{timepoint} column in MNE's physical-examination table distinguishes admission from discharge examinations. This allows us to exclude findings that were available only at discharge, i.e., after the diagnosis had been made (Figure~\ref{fig:intro}).
As shown in Table~\ref{tab:case-study-results}, accuracy matches or improves for every model.
Thus, MNE reproduces the original benchmark context through simple table queries, substantially lowering the curation effort required to construct the benchmark.

\subsection{Agentic CDM Benchmark}
\label{sec:case-study:extensions}

We use MNE's structured representation to develop an agentic CDM benchmark in which agents perform clinical actions to reach a diagnosis (Figure~\ref{fig:benchmark}).
For example, the agent can read a section of the patient history or request a physical examination of a specific body part. This level of granularity was not possible in the original MIMIC-CDM benchmark, where the corresponding tools return entire verbatim text blocks of the whole history or the whole examination notes. The minimal curation effort also allows us to construct a larger benchmark with 10{,}000 cases across 19 pathologies, compared with four pathologies and 2{,}400 cases in MIMIC-CDM.

We create five agent tools representing a subset of real-world clinical actions: reviewing patient history, performing physical examinations, ordering laboratory tests, and requesting imaging or other clinical studies (Table~\ref{tab:cdmv2-tools}).
Agents specify the target of each action, such as examining the abdomen, ordering a troponin test, or requesting a chest X-ray.
In clinical practice, a human doctor would make those requests and receive real findings.
Here, the agent makes the requests through tool calls, and recorded MIMIC and MNE data simulate the findings.
If the information is unavailable, the tool returns a message such as \enquote{The lab cannot perform the troponin test for this patient}.
Overall, the model must decide which clinical action to take next based on the findings obtained so far, determine when it has sufficient evidence, and interpret that evidence to reach a diagnosis.

To accurately simulate patient cases, admissions need sufficient diagnostic evidence.
For example, diagnosing myocardial infarction may require a recorded electrocardiogram (ECG).
MNE enables precise case selection through structured fields, allowing us, e.g., to require an ECG, specific examination systems, or complete vital signs.
A clinician first selected pathologies that could be diagnosed using the available tools, after which 10{,}000 admissions were automatically selected using filters over the MIMIC and MNE tables.

Overall, MNE simplified case filtering and data curation while providing greater flexibility in how information is exposed to the agent.

\paragraph{Results.}
Table~\ref{tab:case-study-results} reports diagnosis accuracy on our benchmark.
The larger models reached 82--88\% accuracy, while the small Qwen3.5 2B only achieved 51.3\%, showing that the agentic benchmark is more challenging for most models than the full-information benchmark.
Reasoning reduced accuracy for Qwen3.6 and improved it for Gemini 3.1 Flash Lite, in line with prior findings that chain-of-thought does not reliably improve clinical performance~\citep{mccoy2025Cot,wu2025chainthoughtfailsclinical}.
See Appendix~\ref{app:cdm-experiments-details} for more details.

%% file: sections/discussion.tex
\section{Discussion}
\label{sec:discussion}

\subsection{Downstream Applications}

Our dataset supports many downstream applications.
Here, we sketch three examples.

\paragraph{Drug-allergy labels and antibiotic prescribing.}
Reported penicillin allergies are associated with higher rates of antibiotic-resistant infections, such as MRSA, and \emph{C.~difficile} infections, partly because they lead to greater use of broad-spectrum alternative antibiotics, even though the reported allergy history is often inaccurate \citep{macy2014penicillin,blumenthal2018penicillin}.
Studying this relationship in MIMIC is difficult because it lacks allergy information. 
Our dataset provides detailed allergy information for nearly 170{,}000 admissions, which can be combined with the broader clinical record to study how reported allergy labels influence antibiotic prescribing and subsequent infection outcomes during hospitalization.

\paragraph{Medication reconciliation.}
\citet{cornish2005discrepancies} showed that more than half of patients admitted on four or more regular medications have at least one unintended discrepancy between their pre-admission medications and their admission orders, and that nearly 40\% of these discrepancies had the potential to cause moderate to severe harm.
MNE provides structured medication lists for ${\sim}300$k admissions.
These lists can be compared with inpatient prescription records to identify medication discrepancies at the cohort scale, enabling large-scale studies of medication reconciliation and polypharmacy (a patient taking multiple medications at the same time).

\paragraph{Risk adjustment with documented function.}
The Hospital Frailty Risk Score estimates patients' vulnerability to poor health outcomes by assessing their reduced ability to recover from illness or stress, using ICD-10 codes because functional status is often unavailable in routine administrative data~\citep{gilbert2018frailty}. 
Such code-based measures can be affected by incomplete or inaccurate coding \citep{omalley2005icd}.
MNE instead provides directly documented information on patients' functional and mental status, including detailed medical history. 
These fields can be used in outcome models and to assess code-based frailty measures.

\subsection{Clinical Text Extraction with LLMs}

Extracting structured information from clinical notes at the scale of MIMIC-IV surfaced several practical lessons that are likely common to similar efforts.

\paragraph{Schema design.}
Most of our design effort went into defining categories that cleanly partition the clinical notes without being overly broad or narrow. We refined categories iteratively until they covered all common cases; capturing rare edge cases was infeasible. Designing an effective schema required close collaboration with medical experts to decide what is worth extracting and to verify correctness. Attempts to automate schema design with an LLM-in-the-loop did not produce satisfactory results.

\paragraph{Prompt design.}
Current LLMs have strong medical knowledge and can interpret clinical terminology, abbreviations, and reasoning without explicit instruction. The main value of prompt design was to resolve ambiguity: clarifying how borderline cases should be handled, describing dataset-specific formatting patterns, handling documentation errors, or disambiguating fields. However, every rule we added, no matter how precise, gave the model more to think over, making it more likely to overthink simple extractions and overapply exception-handling logic. Finding the right balance was difficult: too few rules left genuine ambiguity unresolved, too many made the model second-guess straightforward cases. This was especially pronounced for reasoning models, which would produce lengthy reasoning traces before their response, even when the correct extraction was obvious. Iterative prompt testing with validation sets was essential.

\paragraph{Automatic validation.}
Even the best LLMs make extraction errors. We used Pydantic validators~\citep{Colvin_Pydantic_Validation_2026} to catch and automatically correct common mistakes like malformed dates, out-of-range values, or type mismatches. Fixing these predictable errors through deterministic postprocessing was more reliable than overly specific prompts.

\paragraph{Managing costs.}
Extracting 16 sections across 300{,}720 admissions cost 1{,}894.61\euro~in API fees (Appendix~\ref{app:cost-analysis}).
We kept costs down through concise prompts and output formats, selecting the cheapest model that met quality requirements, using low thinking budgets for reasoning models, and using rule-based extraction wherever LLMs were not necessary.

\paragraph{Zero data retention (ZDR).}
Clinical notes contain sensitive patient information that should not be shared with third parties through cloud services or APIs.
We require strict ZDR policies that prohibit saving data, e.g., for model training or abuse monitoring.
We considered running local models, but found the hardware requirements infeasible at our scale.

\subsection{Limitations and Future Work}

Our dataset inherits limitations from the source discharge summaries.
The original MIMIC developers de-identified the notes by removing names, contact details, identifying numbers, and explicit dates and times.
As described in Section~\ref{sec:dataset}, the de-identification method also redacted some non-identifying content, resulting in loss of clinically relevant information.
The removal of dates and times limits temporal reconstruction, although MNE uses retained relative expressions such as \enquote{at discharge}, \enquote{three years ago}, and \enquote{since yesterday} to capture temporal context.
More generally, discharge summaries reflect clinicians' retrospective selection of information, may contain text copied from earlier encounters, and may omit distracting observations, competing hypotheses, and uncertainty present during care.

Extraction quality was not systematically compared with open-weight and specialized medical models.
The released pipeline, schemas, prompts, and model configurations support such comparisons in future work.

To preserve clinician wording, we did not fully standardize clinical terminology.
Fields such as diagnoses and procedures retain source wording and abbreviations, so identifying patient cohorts may require terminology mapping or additional filters.
Broader terminology normalization remains future work.

Finally, \textit{Brief Hospital Course}, \textit{History of Present Illness}, and \textit{Discharge Instructions} are too diverse to find a reliable pattern for splitting them into fields. Future work may partition these free-text sections into temporal event chains, following \citet{wang2025mimic22mcts}.

%% file: sections/conclusion.tex
\section{Conclusion}
\label{sec:conclusion}

We presented \textsc{MIMIC-Note-Extract}, a large-scale structured dataset that extends MIMIC's tabular data across 300{,}720 admissions with important clinical information, including diagnoses, home medications, allergies, patient history, and physical examination findings.
Overall, our dataset adds 16 new tables to MIMIC that easily integrate with existing data through admission identifiers.
The information was extracted from the MIMIC-IV-Note discharge summaries using regex rules and LLMs.
Our extraction pipeline was built in close collaboration with clinicians and achieves ${>}92\%$ extraction accuracy across sections, as confirmed by medical expert review.

%% file: sections/appendix.tex
\input{sections/appendix/llm_extraction_examples.tex}
\input{sections/appendix/database_relational_schema.tex}
\input{sections/appendix/pipeline_details.tex}
\input{sections/appendix/cost_analysis.tex}
\input{sections/appendix/full_validation_results}
\input{sections/appendix/mimic_comparison.tex}
\input{sections/appendix/cdm_benchmark.tex}

%% file: sections/appendix/llm_extraction_examples.tex
\section{Prompt and Pydantic Schema Examples}
\subsection{Discharge Diagnosis}
\label{app:example-dx}

The \emph{discharge diagnosis} section illustrates a minimal LLM-assisted extraction.
Listings \ref{lst:dx-model} and \ref{lst:dx-prompt} show the pydantic extraction schema and the system prompt.

\begin{lstlisting}[style=pythoncode, caption={Pydantic model for the \emph{discharge diagnosis} section.}, label={lst:dx-model}]
class DischargeDiagnosis(BaseModel):
    id: int = Field(description="Case ID")
    primary:   list[str] = Field(default_factory=list)
    secondary: list[str] = Field(default_factory=list)
\end{lstlisting}

\begin{lstlisting}[style=promptblock, caption={System prompt for the \emph{discharge diagnosis} section.}, label={lst:dx-prompt}]
You are a medical data extraction specialist. Your task is to extract individual medical diagnoses and conditions from unstructured clinical notes. Categorize as primary or secondary when explicitly labeled, otherwise place all in primary (except for special cases). Preserve exact medical terminology, order and remove enumeration/bullets.

SPECIAL CASES:
- Past medical history (pmh) labeled explicitly with a separate header goes to secondary.
- Preserve relevant '[REDACTED]' placeholders.
- Preserve time identifiers in diagnoses (e.g., 'chronic', 'acute', 'history of').
- Remove duplicates.
- Use context when parsing; a few linebreaks are formatting errors, not semantic breaks.
- Special axis structure: preserve as SINGLE primary diagnosis with linebreaks between axes.
- Clear 'with' descriptions are a single diagnosis (e.g., 'Right-sided rib fractures 8 and 9 with small anterior pneumothorax').
\end{lstlisting}

\subsection{Physical Examination}
\label{app:example-pe}

The \emph{physical examination} section is one of the most complex sections.
Each admission can have multiple exam events (admission, discharge, interim) and most events contain a vital-sign object plus a list of per-system assessments.
Enums are encoded with short three-letter codes to keep the LLM's JSON output compact. Listings~\ref{lst:pe-model} and \ref{lst:pe-prompt} show the extraction model and prompt in abbreviated form.

\begin{lstlisting}[style=pythoncode, caption={Abbreviated Pydantic model for the \emph{physical examination} section.}, label={lst:pe-model}]
class VitalSigns(BaseModel):
    temperature:              float | "[REDACTED]" | None
    heart_rate:               int   | "[REDACTED]" | None
    blood_pressure_systolic:  int   | "[REDACTED]" | None
    blood_pressure_diastolic: int   | "[REDACTED]" | None
    respiratory_rate:         int   | "[REDACTED]" | None
    o2_saturation:            int   | "[REDACTED]" | None
    weight:                   Weight | None
    respiratory_support:      OxygenSupport | None

class SystemAssessment(BaseModel):
    system:            SystemName    # GEN, HNT, CVR, RSP, GIT, EXT, NEU, SKN, PSY, GUR, ACC
    status:            ExamStatus    # NRM, ABN, NAS
    abnormal_findings: list[str]

class PhysicalExamEvent(BaseModel):
    timepoint:   Timepoint           # ADM, DSC, INT, UNK
    vital_signs: VitalSigns | None
    systems:     list[SystemAssessment]

class PhysicalExams(BaseModel):
    id:    int
    exams: list[PhysicalExamEvent]
\end{lstlisting}

\begin{lstlisting}[style=promptblock, caption={System prompt for the \emph{physical examination} section.}, label={lst:pe-prompt}]
Role: Medical Data Extraction Specialist
Objective: Extract physical examination data from clinical discharge reports into a structured format while adhering to strict logic for data integrity and medical accuracy.

I. Data Integrity & Constraints
1. System Uniqueness: For any single timepoint (e.g., Admission), a body system must appear only once. If the source contains multiple observations for the same system, synthesize them into a single comprehensive entry.
2. Omission Policy: Only include systems explicitly mentioned in the text. Do not generate entries for systems that were not documented or assessed.
3. Timepoint Separation: Maintain strict separation between distinct exam events (e.g., Admission vs. Discharge). Do not merge findings across different clinical timeframes.

II. The Mutual Exclusivity Rule
Every entry must strictly follow status-finding dependencies.
- Normal Status: "abnormal_findings" must be empty.
- Abnormal Status: if any specific abnormal finding, observation, or indwelling device is extracted, status is "Abnormal".
- Access Systems: presence of any indwelling device (IV, Foley, Chest Tube) dictates status "Abnormal", with the device type listed in "abnormal_findings".
- Negative Indicators: phrases like "No wheezing", "Clear", "Supple", "S1/S2", "Soft", "Non-tender", "Non-distended", "Intact", "Alert", "Oriented" map to Normal status with no abnormal findings.

III. Vital Sign Extraction & Clustering
- Aggressive Re-scan: Vital signs (BP, HR, RR, Temp, SpO2) often appear in clusters. If one metric is identified (e.g., via a "/" for BP or "%" for SpO2), re-scan the surrounding text for unlabeled values.
- Physiological Range Mapping: Use context to map unlabeled numbers. BP is "120/80"; a value of 98 in a cluster is likely saturation or temperature; use standard physiological ranges to differentiate heart rate from respiratory rate.

IV. Clinical Data Transformation
- Terminology Standardization: Translate descriptive observations into standard medical terminology in "abnormal_findings" (e.g., "Rapid breathing" -> Tachypnea, "Redness and swelling" -> Erythema and Edema).
- Synthesized Descriptions: Observations should be concise but clinically accurate.
\end{lstlisting}

%% file: sections/appendix/database_relational_schema.tex
\section{Complete Relational Database Schema}
\label{app:full-schema}

We present the complete database schema of \textsc{MIMIC-Note-Extract} in Table~\ref{tab:full_schema_appendix}.
This schema includes all 16 tables, including administrative hub tables and narrative text tables.
To construct the final relational database, we flatten the Pydantic extraction models. 
Simple fields map directly to standard SQL types, and categorical values convert to SQL enums, while complex polymorphic structures (such as varied medication frequencies) are serialized into \texttt{JSONB} columns to prevent excessive table sparsity. 

\begin{table}[tb]
	\centering
	\footnotesize
	\caption{All 16 tables in the \textsc{MIMIC-Note-Extract} schema. \enquote{Cardinality} refers
		to rows per admission.}
	\label{tab:full_schema_appendix}
	\begin{tabular}{llp{7.5cm}}
		\toprule
		Table                               & Cardinality    & Key fields                                                                                                                         \\
		\midrule
		\multicolumn{3}{l}{\emph{Hub}}                                                                                                                                                            \\
		\texttt{admissions}                 & 1              & \texttt{subject\_id}, \texttt{hadm\_id}, \texttt{note\_id}                                                                         \\
		\midrule
		\multicolumn{3}{l}{\emph{Single-row attributes}}                                                                                                                                          \\
		\texttt{gender}                     & 0\textendash 1 & enum                                                                                                                               \\
		\texttt{services}                   & 0\textendash 1 & enum (20 services)                                                                                                                 \\
		\texttt{discharge\_disposition}     & 0\textendash 1 & enum                                                                                                                               \\
		\texttt{discharge\_condition}       & 0\textendash 1 & 14 enums + 2 booleans                                                                                                              \\
		\midrule
		\multicolumn{3}{l}{\emph{Multi-row clinical lists}}                                                                                                                                       \\
		\texttt{allergies}                  & 0\textendash n & \texttt{title}, \texttt{title\_std}                                                                                                \\
		\texttt{chief\_complaint}           & 0\textendash n & \texttt{category} (enum), \texttt{text}                                                                                            \\
		\texttt{discharge\_diagnosis}       & 0\textendash n & \texttt{title}, \texttt{is\_primary}                                                                                               \\
		\texttt{major\_procedures}          & 0\textendash n & \texttt{title}, \texttt{is\_previous}                                                                                              \\
		\texttt{past\_medical\_history}     & 0\textendash n & \texttt{concept\_name}, \texttt{category}, \texttt{temporal\_status}, \texttt{subject} (patient/family), \texttt{family\_relation} \\
		\texttt{medications\_on\_admission} & 0\textendash n & name, presentation\_form, dose (value + unit), route, duration (value + unit), strength and frequency (JSONB), PRN/hold flags      \\
		\texttt{discharge\_medications}     & 0\textendash n & name, presentation\_form, dose (value + unit), route, duration (value + unit), strength and frequency (JSONB), PRN/hold flags      \\
		\midrule
		\multicolumn{3}{l}{\emph{Per-timepoint structured exam}}                                                                                                                                  \\
		\texttt{physical\_exam\_vitals}     & 0\textendash 4 & \texttt{timepoint} (enum), temperature, heart rate, blood pressure, respiratory rate, O$_2$ saturation, weight, O$_2$ support      \\
		\texttt{physical\_exam\_systems}    & 0\textendash n & \texttt{timepoint} (enum), \texttt{system} (enum), \texttt{status} (enum), \texttt{abnormal\_findings}                             \\
		\midrule
		\multicolumn{3}{l}{\emph{Subsection-typed list}}                                                                                                                                          \\
		\texttt{pertinent\_results}         & 0\textendash n & content type (enum), line range, verbatim text passage, per-type enum arrays, topic label (clinical notes only)                                           \\
		\midrule
		\multicolumn{3}{l}{\emph{Raw text}}                                                                                                                                                       \\
		\texttt{discharge\_sections}        & 1              & one text column per section (16)                                                                                                   \\
		\bottomrule
	\end{tabular}
\end{table}

%% file: sections/appendix/pipeline_details.tex
\section{Pipeline Implementation Details}
\label{app:pipeline-details}

This section provides technical implementation details for the pipeline's section splitting, extraction, and validation stages.

\paragraph{Structural parsing}
Twenty-four distinct clinical sections were identified as appearing in consistent relative order across the MIMIC-IV-Note discharge summaries.
Eight of these sections (e.g., \textit{Name}, \textit{Admission Date}, \textit{Date of Birth}) are entirely redacted by upstream de-identification and are consequently dropped from the cohort.
Deterministic mappings were established between common header variants and canonical section names.
The relative ordering of sections was utilized to automatically resolve all but 5{,}461 ambiguous headers that could not be parsed automatically due to upstream de-identification (e.g., distinguishing between \textit{Discharge Condition} and \textit{Discharge Disposition} given \enquote{Discharge \_\_\_}).
Those cases were manually resolved by inspecting their content.

\paragraph{Hybrid extraction heuristics}
We use section-specific deterministic rules to reduce computational cost, with a fallback to an LLM only when the complexity exceeds the capabilities of regular expressions.
In the \textit{Allergies}, \textit{Chief Complaint}, \textit{Discharge Diagnosis}, and \textit{Major Procedure} sections, the LLM is bypassed for simple cases: single-line entries that match a curated list of terms or restricted character sets.
In more detailed sections, such as \textit{Discharge Condition}, success relies on a \enquote{zero-remainder} policy. Status indicators are extracted and validated using enums. These indicators are then removed from the original text. If any residual content remains, the entire section is processed by the LLM instead.

\paragraph{Validation and post-processing}
Throughout the pipeline, section-specific Pydantic validators enforce structural and clinical consistency on the extracted data.
We showcase two sections here as an example.
For the \textit{Pertinent Results} section, validators verify that extracted document coordinates represent non-overlapping, chronologically ordered line ranges to maintain the original document's sequence.
In the \textit{Physical Examination} section, more complex logic is applied: validators ensure clinical consistency by flagging assessments marked as \enquote{Normal} that contain abnormal findings, while automatically merging fragmented extractions that share the same clinical timepoint.
Global post-processing includes cleaning whitespace, deduplicating redundant clinical list items, and standardizing clinical units, such as normalizing temperature values to Fahrenheit.
Additionally, the pipeline identifies physiological outliers for researcher review while preserving all verbatim values to maintain a faithful representation of the original clinical text.

%% file: sections/appendix/cost_analysis.tex
\section{Extraction Cost Analysis}
\label{app:cost-analysis}

\subsection{Exploiting API Pricing Asymmetries with Compressed Output Schemas}
\label{app:cost-optimization}

Because the LLM generates a structured JSON response for each submitted sample, structural elements such as field names and enum values are repeated in every output. While these keys are strictly necessary for parsing, they are essentially discarded after extraction---yet we still have to pay for the tokens to generate them. At a cohort size of over 300,000 samples, this repetitive JSON overhead becomes a significant driver of total costs.

This presents an excellent opportunity for optimization by exploiting the asymmetric pricing model of LLM APIs, where output tokens are much more expensive than input tokens (e.g., Gemini 3.1 Flash-Lite pricing of \$1.50 per 1M output tokens versus \$0.25 per 1M input tokens). Input tokens---i.e., the static system instructions and schemas repeated across \textit{all} samples---are cheaper still at a rate of \$0.025 per 1M tokens, representing a 60-fold cost reduction compared to output tokens. Consequently, minimizing the number of output tokens is a great lever to save costs.

To exploit this asymmetry, we designed our Pydantic schemas to use short, abbreviated field aliases and enum codes during model inference (e.g., mapping \texttt{subject} $\rightarrow$ \texttt{s}, and \texttt{PATIENT} $\rightarrow$ \texttt{P}). These brief aliases are resolved back to their canonical, human-readable forms downstream to maintain usability for subsequent tasks and human validation.

\newcommand{\codehl}[1]{\textcolor{red}{\textbf{#1}}}

To validate the efficiency of this approach, we conducted a comparative analysis of 20 samples from the \textit{Past Medical History} section using Gemini 3.1 Flash-Lite for the cost analysis.
The full Pydantic models for both the original verbose schema and the optimized version are presented in Listings~\ref{lst:orig_schema} and~\ref{lst:opt_schema}, respectively.
The experimental results demonstrate a 9.87\% reduction in total output tokens compared to the verbose baseline.
This resulted in a 5.99\% reduction in total execution cost under standard pricing.
\footnote{We note that the efficiency of this optimization strategy is highly dependent on the nature of the targeted section. The cost reduction is maximized when the JSON key length is high relative to the extracted value length.}

\paragraph{Implementation note.} This optimization strategy was developed midway through the project, meaning it is not used across all sections. For sections that had already been processed in earlier stages of development, we elected to retain the original verbose schemas rather than incurring the additional cost of re-running the full extraction pipeline with the optimized Pydantic schemas.

	\begin{minipage}[t]{0.48\textwidth}
		\begin{lstlisting}[style=pythoncode, caption={Original verbose Pydantic schema.}, label={lst:orig_schema}]
class SubjectType(str, Enum):
    PATIENT = "PATIENT"
    FAMILY = "FAMILY"

class CategoryType(str, Enum):
    CONDITION = "CONDITION"
    INTERVENTION = "INTERVENTION"
    SUBSTANCE = "SUBSTANCE"
    FUNCTIONAL_STATUS = "FUNCTIONAL_STATUS"

class TemporalStatus(str, Enum):
    CURRENT = "CURRENT"
    PAST = "PAST"

class MedicalHistoryItem(BaseModel):
    subject: SubjectType
    family_relation: str | None = None
    concept_name: str
    category: CategoryType
    temporal_status: TemporalStatus | None = None
		\end{lstlisting}
	\end{minipage}
	\hfill
	\begin{minipage}[t]{0.48\textwidth}
		\begin{lstlisting}[style=pythoncode, caption={Optimized Pydantic schema with 1-token aliases. \textcolor{red}{Red text} indicates the fields and values modified to minimize output token consumption.}, label={lst:opt_schema}, escapechar=\%]
class SubjectType(str, Enum):
    % \codehl{P} % = "%\codehl{P}%"
    % \codehl{F} % = "%\codehl{F}%"

class CategoryType(str, Enum):
    % \codehl{C} % = "%\codehl{C}%"
    % \codehl{I} % = "%\codehl{I}%"
    % \codehl{S} % = "%\codehl{S}%"
    % \codehl{FS} % = "%\codehl{FS}%"

class TemporalStatus(str, Enum):
    % \codehl{U} % = "%\codehl{U}%"
    % \codehl{T} % = "%\codehl{T}%"

class MedicalHistoryItem(BaseModel):
    % \codehl{s} %: SubjectType
    % \codehl{r} %: str | None = None
    % \codehl{n} %: str
    % \codehl{cat} %: CategoryType
    % \codehl{tmp} %: TemporalStatus | None = None
		\end{lstlisting}
	\end{minipage}

\subsection{Detailed Cost and Savings Calculations}
\label{app:cost-calculations}

A detailed breakdown of the extraction costs and savings afforded by our hybrid, rule-first extraction strategy is presented.
To calculate these figures, a baseline cost for each clinical section, assuming a fully LLM-driven pipeline, is first estimated.
This baseline is established by calculating the average per-case cost from representative pilot runs using Gemini pricing models.
This unit cost is then multiplied by the total number of cases in the cohort to project the baseline expense.

The baseline is compared with actual production expenditures, and the resulting net savings are shown in Table~\ref{tab:unified_costs_appendix}.
The table structure is defined as follows:
\begin{itemize}
    \item \textbf{LLM Baseline:} The projected total cost of extracting the entire section using only LLMs.
    \item \textbf{Production Cost:} The actual cost incurred using our optimized pipeline, where deterministic rules bypassed LLM calls for structured cases.
    \item \textbf{Optimized Savings:} The savings achieved by our rule-first strategy.
\end{itemize}

\begin{table}[tb]
	\centering
	\footnotesize
	\setlength{\tabcolsep}{4pt}
	\renewcommand{\arraystretch}{1.15}
	\caption{Production extraction methodology, LLM model, API expenditures, and rule-first optimization savings by clinical section.}
	\label{tab:unified_costs_appendix}
	\begin{tabular*}{\linewidth}{@{\extracolsep{\fill}}lllrrr@{}}
		\toprule
		\shortstack[l]{\textbf{Clinical}\\\textbf{Section}} & \shortstack[l]{\textbf{Operational}\\\textbf{Mode}} & \shortstack[l]{\textbf{Production}\\\textbf{Model}} & \shortstack[r]{\textbf{LLM}\\\textbf{Baseline (\euro)}} & \shortstack[r]{\textbf{Production}\\\textbf{Cost (\euro)}} & \shortstack[r]{\textbf{Optimized}\\\textbf{Savings (\euro)}} \\
		\midrule
		\multicolumn{6}{l}{\emph{LLM-Only Extraction}} \\
		Past Medical History       & LLM                       & Gemini 3.1 Flash-Lite         & 138.07                        & 138.07                           & --- \\
		Medications on Admission   & LLM                       & Gemini 3.1 Flash-Lite         & 487.00                        & 487.00                           & --- \\
		Physical Examination       & LLM                       & Gemini 3.1 Flash-Lite         & 345.00                        & 345.00                           & --- \\
		Pertinent Results          & LLM                       & Gemini 3.1 Flash-Lite         & 179.00                        & 179.00                           & --- \\
		Discharge Medications      & LLM                       & Gemini 3.1 Flash-Lite         & 581.00                        & 581.00                           & --- \\
		\midrule
		\multicolumn{6}{l}{\emph{Hybrid (Rule-First with LLM Fallback)}} \\
		Chief Complaint            & Hybrid (61.2\% Auto)      & Gemini 2.5 Flash              & 62.78                         & 24.36                            & 38.42 \\
		Allergies                  & Hybrid (99.1\% Auto)      & Gemini 2.5 Flash              & 124.90                        & 1.07                             & 123.83 \\
		Major Procedures           & Hybrid (20.0\% Auto)      & Gemini 2.5 Flash              & 53.42                         & 42.75                            & 10.67 \\
		Discharge Diagnosis        & Hybrid (27.6\% Auto)      & Gemini 2.5 Flash              & 63.68                         & 46.11                            & 17.57 \\
		Discharge Condition        & Hybrid (72.2\% Auto)      & Gemini 3.1 Flash-Lite         & 181.00                        & 50.25                            & 130.75 \\
		\midrule
		\multicolumn{6}{l}{\emph{Deterministic (Rule / Verbatim)}} \\
		Gender                     & Rule-Based                & ---                           & 0.00                          & 0.00                             & --- \\
		Services                   & Rule-Based                & ---                           & 0.00                          & 0.00                             & --- \\
		Discharge Disposition      & Rule-Based                & ---                           & 0.00                          & 0.00                             & --- \\
		History of Present Illness & Verbatim                  & ---                           & 0.00                          & 0.00                             & --- \\
		Brief Hospital Course      & Verbatim                  & ---                           & 0.00                          & 0.00                             & --- \\
		Discharge Instructions     & Verbatim                  & ---                           & 0.00                          & 0.00                             & --- \\
		\midrule
		\textbf{Aggregate Total}   &                           & Mixed                         & \textbf{2{,}215.85}             & \textbf{1{,}894.61}                & \textbf{321.24} \\
		\bottomrule
	\end{tabular*}
\end{table}

%% file: sections/appendix/full_validation_results.tex
\clearpage
\section{Full Validation Results}
\label{app:validation-full}

This section provides the full details of the manual validation, including the error taxonomy, concrete examples, metric definitions, and the complete performance breakdown.

\paragraph{Error taxonomy} \nopagebreak
\begin{description} 
  \item[Ambiguous] The source text does not unambiguously support either approval or rejection (e.g., shorthand that could be interpreted in multiple valid ways).
  \item[Fabrication] The extracted item is not supported by the source text.
  \item[Omission] An item present in the source text is missing from the extracted set.
  \item[Clinically Altered Meaning] The extracted item is textually close to the source but semantically different in a way that would affect a clinical reader.
  \item[Incorrect Attribute / Classification] Structured attribute fields (e.g., dosage or category) are incorrect.
  \item[Incorrect Boundaries] The detected line boundaries are incorrect. Exclusive to the \enquote{Pertinent Results} section.
\end{description}

\paragraph{Concrete examples}
Table~\ref{tab:error-examples} illustrates the clinician-assigned labels with source excerpts and the corresponding extracted fields.
We selected illustrative examples, one for each category.
Each explanation follows the clinicians' annotations.

\begin{table}[htbp]
  \centering
  \footnotesize
  \caption{Examples of clinician-assigned error and ambiguity labels.
  Bold text highlights the source phrases and extracted values relevant to each judgment; field names are set in monospace.
  Source excerpts retain the original wording, with line breaks represented by spaces or slashes except in the boundary example; extracted fields are abbreviated for display.
  Boundary line numbers are zero-based and inclusive.}
  \label{tab:error-examples}
  \input{tables/error_examples.tex}
\end{table}

An overview of extraction errors and ambiguous judgments across all categories is detailed in Table~\ref{tab:error-breakdown}.
Validation highlights that omissions (231 errors) and incorrect attributes (157 errors) are the predominant error modes, localized mainly within \enquote{Physical Examination} and \enquote{Past Medical History}.
In contrast, critical failures such as fabrications (14 total) and clinically altered meanings (42 total) are rare across the 13{,}744 validated items.

Accuracy is calculated using an ambiguity-excluded mapping. Because an ambiguous judgment indicates that the source text does not support a reliable decision about correctness, we exclude it from the accuracy denominators rather than treating it as either correct or incorrect.
At the item level, each remaining extracted item contributes one decision, and recorded omissions are added to the denominator as incorrect missing items.
Precision, recall, and F1-score use the same mapping, treating omissions as false negatives and fabrications or semantic alterations as false positives.
At the case level, we evaluate the remaining item decisions and exclude cases with no evaluable items after removing ambiguous judgments; a case with no items and no recorded omissions is scored as correct by convention.

Judgments are clustered within cases: the 50 dual-annotated cases contribute two judgments each, and the items of a case are reviewed together rather than drawn independently.
We therefore take the case as the unit of analysis, letting each case contribute the mean of the judgments made about it, so that a disagreement between two annotators enters as 0.5.
Margins of error come from a cluster bootstrap that resamples the 171 cases with replacement over 10{,}000 replicates and reports the 2.5th and 97.5th percentiles, which anchors the effective sample size to the independent cases rather than to the judgments made about them.

The item-level accuracy column is the Jaccard index $\mathit{TP} / (\mathit{TP} + \mathit{FP} + \mathit{FN})$ rather than conventional accuracy.
A true negative would be an item correctly not extracted, which carries no meaning for a task whose objective is to extract everything present in the source text, so the taxonomy has no such category and none can enter a denominator.
The case-level columns are macro-averaged over cases while the item-level columns are micro-averaged over items, and a case with no items and no recorded omissions is scored as correct by convention.

To assess inter-annotator agreement, both raw agreement and Gwet's AC1 are reported using a tolerant mapping that treats \textit{Approve} and \textit{Ambiguous} as the same label. This measures agreement on substantive extraction correctness without penalizing differences in annotators' thresholds for assigning the ambiguity label.
Raw agreement can be misleadingly high in datasets with high-prevalence categories (as observed here).
Gwet's AC1 is more robust to this effect and provides a more stable estimate of agreement beyond chance.
It is reported at the item level only; a case-level label is the full pattern of item labels across the case, which leaves nearly every case in a category of its own and reduces the chance correction to nothing.

Error counts are reported per annotator judgment rather than per case, so an error flagged independently by both annotators of a dual-annotated case contributes twice.
The \enquote{Judged Items} column is the matching denominator, leaving the relative figures unaffected.

The gap between case-level and item-level accuracy (Tables~\ref{tab:validation-case-performance} and~\ref{tab:validation-item-performance}) is expected in multi-item sections: a single disputed medication or history item can cause the entire case to fail. 
This is most evident in complex sections such as past medical history or medications. 
Item-level quality is treated as the primary estimate for cohort-level reuse, while case-level accuracy provides a conservative lower bound.

The agreement gap in the medication sections stems from how annotators handled incomplete dosage and strength details in the raw notes.
When the source text omitted medication strengths, one clinician routinely flagged the extraction as \enquote{Ambiguous} while others marked it as \enquote{Approved}.
While this difference in labeling threshold lowered the exact raw agreement, both annotators agreed that the LLM successfully extracted everything present in the source.

Tables~\ref{tab:validation-case-performance} and~\ref{tab:validation-item-performance} report the complete per-section validation breakdown of model performance at the case and item levels, while Table~\ref{tab:validation-iaa} provides the detailed inter-annotator agreement results. 
Table~\ref{tab:error-breakdown} provides the counts and percentages of extraction errors and ambiguous judgments across all clinical sections.

\begin{table}[ht]
  \centering
  \scriptsize
  \setlength{\tabcolsep}{5pt}
  \caption{Counts and percentages of extraction errors and ambiguous judgments by clinical section.
  Percentages are calculated relative to all judged items within each section.}
  \label{tab:error-breakdown}
  \input{tables/error_breakdown.tex}
\end{table}

\begin{table}[tb]
  \centering
  \scriptsize
  \caption{Model extraction performance by clinical section at the \emph{case} level (with 95\% confidence margins of error).}
  \label{tab:validation-case-performance}
  \input{tables/case_performance.tex}
\end{table}

\begin{table}[tb]
  \centering
  \scriptsize
  \caption{Model extraction performance by clinical section at the \emph{item} level (with 95\% confidence margins of error).}
  \label{tab:validation-item-performance}
  \input{tables/item_performance.tex}
\end{table}

\begin{table}[tb]
  \centering
  \scriptsize
  \caption{Inter-Annotator Agreement (Raw IAA and Gwet's AC1) by clinical section.}
  \label{tab:validation-iaa}
  \input{tables/iaa_agreement.tex}
\end{table}

%% file: tables/error_examples.tex
\begingroup
\setlength{\tabcolsep}{5pt}
\renewcommand{\arraystretch}{1.18}
\begin{tabular}{@{}>{\raggedright\arraybackslash}p{0.15\dimexpr\linewidth-30pt\relax}>{\raggedright\arraybackslash}p{0.30\dimexpr\linewidth-30pt\relax}>{\raggedright\arraybackslash}p{0.23\dimexpr\linewidth-30pt\relax}>{\raggedright\arraybackslash}p{0.32\dimexpr\linewidth-30pt\relax}@{}}
  \toprule
  \textbf{Category / section} & \textbf{Source excerpt} & \textbf{Extracted fields} & \textbf{Error explanation} \\
  \midrule
  Ambiguous\par\emph{Medications on admission}
  & Simethicone \textbf{40-80 mg} PO QID:PRN abd pain
  & \texttt{strength}: range \textbf{40--80 mg}\par
    \texttt{dose}: \textbf{80 mg}
  & The source gives a 40--80 mg range without separately specifying strength and administered dose.
    The extraction records both a strength range and a single 80 mg dose. \\
  \rowrule
  Fabrication\par\emph{Past medical history}
  & R THA / L THA / \textbf{Stents x2}
  & \texttt{concept\_name}: \textbf{Coronary Artery} Stents
  & The extraction adds a coronary location to the stents, although their anatomical location is not specified in the source. \\
  \rowrule
  Omission\par\emph{Medications on admission}
  & 18. \textbf{Insulin SC} / Sliding Scale\par
    Insulin SC Sliding Scale using HUM Insulin
  & \texttt{medications}: \textbf{no insulin entry}
  & Sliding-scale insulin is explicitly listed in the source but is absent from the complete extracted medication list. \\
  \rowrule
  Clinically altered meaning\par\emph{Discharge medications}
  & Calcium 500 + D (calcium carbonate-vitamin D3) 500 mg\textbf{(1,250mg)} -400 unit oral DAILY
  & \texttt{strength}: [500 mg, 400 units]\par
    \texttt{dose}: \textbf{1{,}250 mg}
  & The parenthetical 1{,}250 mg quantity is incorrectly interpreted as the administered dose. \\
  \rowrule
  Incorrect attribute / classification\par\emph{Discharge medications}
  & furosemide 40 mg Tablet \_\_\_: One (1) Tablet PO \textbf{ONCE (Once) for 1 doses}.
  & \texttt{name}: furosemide\par
    \texttt{frequency}: \textbf{1 per DAY}\par
    \texttt{duration}: null
  & A one-time dose is incorrectly represented as daily administration without a duration limit. \\
  \rowrule
  Incorrect boundaries\par\emph{Pertinent results}
  & \texttt{19:} collection.\par
    \texttt{20:} \textbf{Findings were concerning for early or partial small bowel}\par
    \texttt{21:} \textbf{obstruction. The NG tube was kept in place, and medications were}
  & \texttt{start\_line}: 6\par
    \texttt{end\_line}: \textbf{19}\par\smallskip
    Clinician span: \textbf{6--21}
  & The extracted span ends at line 19 instead of line 21, omitting the following finding of suspected small-bowel obstruction. \\
  \bottomrule
\end{tabular}
\endgroup

%% file: tables/error_breakdown.tex
  \begin{tabular}{lccccccc}
    \toprule
    \textbf{Section} & \textbf{Judged Items} & \textbf{Omiss.} & \textbf{Attr.} & \textbf{Alter.} & \textbf{Fabr.} & \textbf{Bound.} & \textbf{Ambig.} \\
    \midrule
    Allergies & 1,488 & 0 (0.0\%) & 0 (0.0\%) & 1 (0.1\%) & 2 (0.1\%) & 0 (0.0\%) & 6 (0.4\%) \\
    Chief Complaint & 222 & 1 (0.5\%) & 4 (1.8\%) & 0 (0.0\%) & 0 (0.0\%) & 0 (0.0\%) & 3 (1.4\%) \\
    Discharge Condition & 222 & 4 (1.8\%) & 0 (0.0\%) & 0 (0.0\%) & 1 (0.5\%) & 0 (0.0\%) & 1 (0.5\%) \\
    Discharge Diagnosis & 222 & 1 (0.5\%) & 5 (2.3\%) & 0 (0.0\%) & 0 (0.0\%) & 0 (0.0\%) & 0 (0.0\%) \\
    Major Procedures & 222 & 1 (0.5\%) & 1 (0.5\%) & 2 (0.9\%) & 0 (0.0\%) & 0 (0.0\%) & 0 (0.0\%) \\
    Past Medical History & 2,761 & 66 (2.4\%) & 46 (1.7\%) & 26 (0.9\%) & 3 (0.1\%) & 0 (0.0\%) & 36 (1.3\%) \\
    Physical Examination & 3,715 & 103 (2.8\%) & 79 (2.1\%) & 3 (0.1\%) & 0 (0.0\%) & 0 (0.0\%) & 6 (0.2\%) \\
    Pertinent Results & 564 & 23 (4.1\%) & 4 (0.7\%) & 6 (1.1\%) & 0 (0.0\%) & 10 (1.8\%) & 1 (0.2\%) \\
    Discharge Medications & 2,281 & 21 (0.9\%) & 13 (0.6\%) & 4 (0.2\%) & 7 (0.3\%) & 0 (0.0\%) & 551 (24.2\%) \\
    Medications on Admission & 2,047 & 11 (0.5\%) & 5 (0.2\%) & 0 (0.0\%) & 1 (0.0\%) & 0 (0.0\%) & 388 (19.0\%) \\
    \midrule
    \textbf{Total} & \textbf{13,744} & \textbf{231 (1.7\%)} & \textbf{157 (1.1\%)} & \textbf{42 (0.3\%)} & \textbf{14 (0.1\%)} & \textbf{10 (0.1\%)} & \textbf{992 (7.2\%)} \\
    \bottomrule
  \end{tabular}

%% file: tables/case_performance.tex
  \begin{tabular}{lcccc}
    \toprule
    & \multicolumn{4}{c}{\textbf{\textsc{Ambiguity-excluded Case-level (\%)}}} \\
    \cmidrule(lr){2-5}
    \textbf{Section} & \textbf{Acc.} & \textbf{Prec.} & \textbf{Rec.} & \textbf{F1} \\
    \midrule
    Allergies                  & 98.5 $\pm$ 2.06 & 99.6 & 100.0 & 99.8 \\
    Chief Complaint            & 97.6 $\pm$ 2.37 & 97.9 & 97.8 & 97.8 \\
    Discharge Condition        & 97.9 $\pm$ 2.35 & 97.9 & 97.9 & 97.9 \\
    Discharge Diagnosis        & 97.1 $\pm$ 2.63 & 97.7 & 97.4 & 97.5 \\
    Major Procedures           & 98.2 $\pm$ 2.05 & 98.5 & 98.4 & 98.4 \\
    Past Medical History       & 72.8 $\pm$ 6.43 & 97.7 & 97.4 & 97.2 \\
    Physical Examination       & 80.7 $\pm$ 5.56 & 97.3 & 95.2 & 96.1 \\
    Pertinent Results          & 89.5 $\pm$ 4.39 & 95.6 & 96.3 & 95.7 \\
    Discharge Medications      & 84.8 $\pm$ 5.15 & 98.4 & 98.9 & 98.4 \\
    Medications on Admission   & 95.4 $\pm$ 2.78 & 99.5 & 99.6 & 99.5 \\
    \bottomrule
  \end{tabular}

%% file: tables/item_performance.tex
  \begin{tabular}{lcccc}
    \toprule
    & \multicolumn{4}{c}{\textbf{\textsc{Ambiguity-excluded Item-level (\%)}}} \\
    \cmidrule(lr){2-5}
    \textbf{Section} & \textbf{Acc.} & \textbf{Prec.} & \textbf{Rec.} & \textbf{F1} \\
    \midrule
    Allergies                  & 99.8 $\pm$ 0.32 & 99.8 & 100.0 & 99.9 \\
    Chief Complaint            & 97.6 $\pm$ 2.37 & 97.9 & 99.7 & 98.8 \\
    Discharge Condition        & 97.4 $\pm$ 3.15 & 99.4 & 97.9 & 98.7 \\
    Discharge Diagnosis        & 97.1 $\pm$ 2.65 & 97.7 & 99.4 & 98.5 \\
    Major Procedures           & 98.3 $\pm$ 2.04 & 98.5 & 99.7 & 99.1 \\
    Past Medical History       & 95.2 $\pm$ 1.48 & 97.5 & 97.5 & 97.5 \\
    Physical Examination       & 95.2 $\pm$ 1.83 & 98.1 & 97.0 & 97.5 \\
    Pertinent Results          & 92.2 $\pm$ 3.81 & 96.2 & 95.7 & 96.0 \\
    Discharge Medications      & 97.2 $\pm$ 1.17 & 98.6 & 98.6 & 98.6 \\
    Medications on Admission   & 99.3 $\pm$ 0.47 & 99.7 & 99.5 & 99.6 \\
    \bottomrule
  \end{tabular}

%% file: tables/iaa_agreement.tex
  \begin{tabular}{lccc}
    \toprule
    \textbf{Section} & \textbf{Case IAA (tolerant)} & \textbf{Item IAA (tolerant)} & \textbf{AC1 (tolerant)} \\
    \midrule
    Allergies                  & 98.0 & 99.8 & 99.8 \\
    Chief Complaint            & 96.0 & 98.0 & 98.0 \\
    Discharge Condition        & 98.0 & 98.0 & 98.0 \\
    Discharge Diagnosis        & 96.0 & 96.0 & 95.8 \\
    Major Procedures           & 96.0 & 98.0 & 98.0 \\
    Past Medical History       & 56.0 & 90.4 & 90.1 \\
    Physical Examination       & 68.0 & 90.9 & 90.4 \\
    Pertinent Results          & 80.0 & 88.7 & 88.4 \\
    Discharge Medications      & 76.0 & 97.1 & 97.1 \\
    Medications on Admission   & 80.0 & 97.3 & 97.2 \\
    \bottomrule
  \end{tabular}

%% file: sections/appendix/mimic_comparison.tex
\section{Details on Comparison to MIMIC Tables}
\label{app:mimic-comparison-details}

In this section, we provide details on the comparison of each MNE section to the existing MIMIC tables.
MIMIC tables span different modules, which differ in both content and cohort.
The hospital-wide \texttt{hosp} module covers all admissions, while the emergency-department (ED) and intensive-care (ICU) modules add finer triage and bedside detail but only for the admissions treated in those units.

\begin{table*}[tb]
  \centering
  \footnotesize
  \caption{Extracted sections grouped by how they relate to MIMIC's existing tables, with the closest related table(s) and the contribution of our extraction.
  Table prefixes: h=hosp, e=ED, i=ICU.}
  \label{tab:mimic-comparison-overview-full}
  \setlength{\tabcolsep}{3pt}%
  \begin{tabular}{>{\raggedright\arraybackslash}p{0.23\linewidth}>{\raggedright\arraybackslash}p{0.28\linewidth}p{0.45\linewidth}}
    \toprule
    \textbf{Section} & \textbf{Related MIMIC table(s)} & \textbf{Contribution} \\
    \midrule
    \multicolumn{3}{l}{\emph{Already exists in MIMIC, but can flag mismatches}} \\
    \rowrule
    Gender & h.patients.gender & Flag mismatches \\
    \rowrule
    Services & h.services.curr\_service & Flag mismatches \\
    \rowrule
    Discharge Disposition & h.admissions.discharge\_location & Flag mismatches; adds ${\sim}44$k missing values \\
    \midrule
    \multicolumn{3}{l}{\emph{Related to MIMIC information, but broader or richer}} \\
    \rowrule
    Discharge Diagnosis & h.diagnoses\_icd & Concise clinician-written diagnoses with primary/secondary structure, preserving note wording instead of dense, generic ICD codes \\
    \rowrule
    Chief Complaint & e.triage.chiefcomplaint & 2$\times$ more admissions; cleaner terms and additional clinical context \\
    \rowrule
    Allergies & h.diagnoses\_icd (allergy codes) & Dedicated structured allergen list covering far more admissions than MIMIC's sparse ICD codes \\
    \rowrule
    Past Medical History & h.diagnoses\_icd (history codes) & 10$\times$ more history rows; contains broader prior conditions, family, substance, and functional history, compared to MIMIC's ICD codes \\
    \rowrule
    Medications on Admission & e.medrecon & $>2\times$ more admissions; more detail by providing dose, route, frequency, PRN, and hold status \\
    \rowrule
    Major Procedures & h.procedures\_icd; h.hcpcsevents & Contains compact clinical procedure label and more surgical details compared to coded ICD/HCPCS billing tables \\
    \midrule
    \multicolumn{3}{l}{\emph{No structured MIMIC counterpart}} \\
    \rowrule
    Discharge Condition & none & Provides structured discharge state, including mental status and functional activity \\
    \rowrule
    Discharge Medications & none & Provides discharge-medication list \\
    \rowrule
    Physical Examination & none & Provides system-level examination findings with normal/abnormal status \\
    \rowrule
    Pertinent Results & none & Provides important clinical results, e.g., radiology interpretations \\
    \bottomrule
  \end{tabular}
\end{table*}

\paragraph{Gender}

The extracted \emph{gender} field maps directly to \texttt{hosp.patients.gender}: both use the values \texttt{M} and \texttt{F}.
99.96\% of documented genders match exactly, leaving 122 admissions with differences.
Manual review of these mismatching admissions shows that most are explained by transgender, gender-identity, or transition-related documentation in the discharge notes.
Thus, MNE's \emph{gender} field can serve as a sanity check for correctness.

\paragraph{Services}

The extracted \emph{services} field corresponds to \texttt{hosp.services.curr\_service}. However, MIMIC stores abbreviated transfer-level service codes rather than the broad labels used in our table.
\texttt{hosp.services} represents service history as timed transfer rows, using transfer time and abbreviated codes such as \texttt{MED}, \texttt{SURG}, and \texttt{NSURG}.
We map these codes to our extraction labels before comparison.
Many mappings are one-to-one, but some MIMIC codes map to the same extracted category (e.g.\ \texttt{MED}, \texttt{CMED}, and \texttt{OMED} to \texttt{MEDICINE}), and a few extracted labels have no direct counterpart.
We test whether the discharge-summary header records the service at discharge by comparing it with the last MIMIC service entry.
Among 300{,}221 comparable admissions, 99.16\% match exactly.
Allowing any mapped service during the admission to count as a match increases the overlap by only 175 cases. 
Thus, while MNE largely contains service information already available in MIMIC, it also provides a useful sanity check by flagging cases with mismatched service assignments.

\paragraph{Discharge disposition}

The extracted \emph{discharge disposition} field contains four broad categories: \texttt{HOME}, \texttt{HOME\_WITH\_SERVICE}, \texttt{EXTENDED\_CARE}, and \texttt{EXPIRED}.
\texttt{hosp.admissions} stores the more detailed \texttt{discharge\_location}, together with death indicators such as \texttt{hospital\_expire\_flag} and \texttt{deathtime}.
We map MIMIC discharge locations to extraction labels where the correspondence is clear, for example \texttt{HOME} to \texttt{HOME}, \texttt{HOME HEALTH CARE} to \texttt{HOME\_WITH\_SERVICE}, and skilled nursing or rehabilitation facilities to \texttt{EXTENDED\_CARE}.
The main remaining unmapped MIMIC destinations are \texttt{HOSPICE} and \texttt{AGAINST ADVICE}, which occur in roughly 5{,}000 admissions and are more specific than our four-category schema.
Among 251{,}208 admissions with a mapped non-missing MIMIC disposition, 96.87\% match the extracted value.

However, the MIMIC table is missing a discharge location for 44{,}209 admissions, which are in MNE.
These cases represent added structured information rather than disagreement with MIMIC. Thus, the extracted field validates well against mapped MIMIC values and fills many missing location values.

\paragraph{Allergies}

MIMIC has no dedicated structured allergy table, but \texttt{hosp.diagnoses\_icd} contains 36 allergy-status billing codes for only 2{,}211 admissions such as \texttt{V140} (\enquote{personal history of allergy to penicillin}), \texttt{V1507} (\enquote{latex allergy}), and broader codes such as \texttt{V1505} (\enquote{Allergy to other foods}).
MNE's allergy table covers 169{,}595 admissions with at least one documented allergen and contains 606{,}069 allergy rows overall.
We compare the overlapping subset by selecting ICD-9 allergy-status codes \texttt{V14*}, \texttt{V1504}, \texttt{V1505}, \texttt{V1507}, \texttt{V1508}, and \texttt{V1509}, and ICD-10 codes \texttt{Z88*} and \texttt{Z910*}.
We then map both sources to broad allergen categories where a correspondence is possible, such as penicillin, sulfonamide, contrast, latex, seafood, antibiotic, analgesic, or narcotic allergy, using substring matching.
Using this approximate category mapping, about 70\% of 2{,}211 admissions with ICD allergy-status codes have a matching extracted allergen category in our table.
Overall, we contribute a new table with detailed structured allergy lists, which are not part of MIMIC's tables.

\paragraph{Chief complaint}

The extracted \emph{chief complaint} field has a partial structured counterpart in \texttt{ed.triage.chiefcomplaint}, but the coverage and semantics differ.
The ED triage table captures the presenting complaint at ED arrival and is only available for ED stays, whereas the discharge summary can contain admission complaints, hospital-stay complaints, and additional clinical context.
MNE's \emph{chief complaint} table covers 296{,}132 admissions, while \texttt{ed.triage.chiefcomplaint} is available for 140{,}466 admissions in the discharge-summary cohort.
There are 137{,}035 admissions with both sources.
For these paired admissions, we concatenate multiple extracted chief complaint rows and multiple ED triage complaints within each admission, normalize both strings by lowercasing, removing non-alphanumeric characters, and normalizing whitespace, and compare the resulting strings.
Normalized exact agreement is 11.0\%, while token-set string similarity at the threshold of 90 reaches 29.2\%.
These low agreement rates reflect that the two fields are related but not interchangeable: ED triage values are short arrival complaints and abbreviations, while discharge-summary chief complaints are often more comprehensive with additional clinical context.
The extracted field therefore adds broader hospital-level chief-complaint coverage rather than duplicating ED triage.

\paragraph{Major procedures}

The extracted \emph{major procedures} field is related to the coded procedure sources in the hosp module (\texttt{procedures\_icd} and \texttt{hcpcsevents}), but these tables use standardized procedure codes for billing and reporting, while our field contains concise clinician-written procedure summaries.
Considering only procedures from the current admission, the extracted major-procedure table covers 146{,}515 admissions, while 175{,}599 admissions in our cohort have ICD procedure codes and 44{,}514 have HCPCS events.
The current-admission extracted procedure subset contains 250{,}504 procedure rows, compared with 564{,}069 ICD or HCPCS rows in the related MIMIC tables for the same cohort.
In practice, the sources are only partially comparable: ICD and HCPCS often encode approach, site, device, material, and billing-specific detail, while the discharge note usually gives a compact clinical procedure label or surgical details.
Overall, \emph{major procedures} provide clinician-written procedure summaries to MIMIC.

\paragraph{Admission and discharge medications}

MIMIC contains several medication-related tables, but they answer different questions.
\texttt{hosp.prescriptions}, \texttt{hosp.pharmacy}, \texttt{hosp.emar}, and \texttt{icu.inputevents} describe inpatient orders, administrations, or ICU inputs, not a reconciled pre-admission home-medication list.
The closest structured counterpart for medications on admission is \texttt{ed.medrecon}, which contains ED medication reconciliation rows but covers only the ED subset.
In our discharge-summary cohort, ed.medrecon is available for 116{,}387 admissions, while our extracted admission-medication table covers 275{,}035 admissions.
There are 112{,}294 admissions with both sources and 162{,}741 admissions with extracted admission medications but no mapped ED medication reconciliation.
For the overlap subset, we first compare medication names with conservative normalized exact matching, which gives a lower-bound agreement of 44.72\% from extracted medications to ED medrecon and 43.61\% in the reverse direction.
We then map each medication name to an RxNorm ingredient key when a mapping is available \citep{nelson2011rxnorm}; unmapped names retain their canonicalized string as fallback.
This increases row-level agreement to 74.25\% from extracted medications to ED medrecon and 72.63\% in the reverse direction.
These values should be interpreted as same-admission agreement with ED medication reconciliation, not as extraction accuracy, because the two sources can differ in timing, completeness, and medication-name granularity.
The extracted admission-medication table also provides structured attributes not exposed by \texttt{ed.medrecon}, including dose, strength, route, frequency, duration, PRN status, and hold instructions.
The extracted discharge-medication table covers 285{,}926 admissions and adds a structured discharge-state medication list.
No standard MIMIC table exposes discharge medications as a single reconciled list.

\paragraph{Discharge diagnosis}

The extracted \emph{discharge diagnoses} are related to \texttt{hosp.diagnoses\_icd}, but the two sources differ in purpose.
Our table captures clinician-written diagnoses from the discharge summary, whereas \texttt{diagnoses\_icd} contains standardized billing-coded diagnoses, including comorbidities, complications, history/status codes, and ICD-specific granularity.
Coverage is high in both sources: 297{,}537 admissions have extracted discharge diagnoses, 300{,}546 have ICD diagnoses, and 297{,}375 have both.
However, ICD is much denser, with 3{,}720{,}215 diagnosis rows and a mean of 12.38 rows per admission, compared with 982{,}748 extracted discharge-diagnosis rows and a mean of 3.30 rows per admission.
We first compare same-admission strings directly against ICD long titles.
Raw exact matching links only 4.32\% of extracted diagnosis rows to an ICD title, and conservative normalization increases this to 15.79\%.
This normalization expands common abbreviations such as HTN to hypertension and removes generic ICD title wording such as \enquote{unspecified}, \enquote{other and unspecified}, and \enquote{not elsewhere classified}.
We then use token-set similarity as an approximate string-matching analysis.
At threshold 80, 44.87\% of extracted diagnosis rows have a sufficiently similar same-admission ICD title.
Thus, fewer than half of the diagnoses in our extracted discharge-diagnosis list can be linked to a single billed ICD diagnosis title using this approximate string comparison.
Manual inspection shows that many non-links are representational differences rather than obvious errors: clinician-written diagnoses may use concise phrases, etiologic combinations, post-procedure context, or problem-level labels that require combining several ICD rows or translating ICD-specific wording.
The ICD table remains the appropriate source when standardized billing codes are required.
Our extracted table contributes a compact, directly interpretable clinical diagnosis list with primary versus secondary structure, preserving the wording and grouping used in the discharge summary rather than requiring users to reconstruct this summary from multiple coded rows.

\paragraph{Discharge condition}

The \emph{discharge condition} fields, such as mental status and functional activity, have no direct structured counterpart in MIMIC hosp, ICU, or ED.
This section therefore adds new structured discharge-state information for nearly the full cohort.

\paragraph{Past medical history}

The closest structured MIMIC counterpart for \emph{past medical history} is the subset of history/status billing codes in \texttt{hosp.diagnoses\_icd}, but this is only an indirect comparison.
Our extracted past-medical-history table contains 3{,}616{,}698 rows across 290{,}212 admissions, including clinician-written prior conditions, family history, prior interventions, substance history, and functional status.
The selected ICD history/status subset contains 373{,}062 rows across 181{,}890 admissions, and both sources are present for 178{,}619 admissions.
These ICD codes mainly represent selected billable history and status concepts, such as prior tobacco use, prior thromboembolism, malignancy history, cardiac bypass or angioplasty status, dialysis status, and implanted devices.
Thus, ICD history/status codes provide useful context, but they are not a direct validation target for the extracted past-medical-history section.
Our table adds substantially broader structured clinical and family-history content.

\paragraph{Physical exam}

The \emph{physical exam} section has no direct structured counterpart in MIMIC.
ED triage, ED vitalsign, and ICU chartevents contain measured vitals or bedside observations, often tracked repeatedly over time and only for ED or ICU subsets.
In contrast, our extracted physical-exam tables cover 294{,}708 admissions and represent note-derived examinations at documented timepoints, mainly admission and discharge.
They contain 434{,}633 vital-sign rows and 4{,}780{,}963 system-level exam rows with normal/abnormal status and abnormal findings.
The system-level clinician assessment is the main contribution, because it is not available as an equivalent structured MIMIC hospital table.
Therefore, the related MIMIC measurement tables are orthogonal rather than comparable.

\paragraph{Pertinent results}

The \emph{pertinent results} table has no direct structured counterpart in MIMIC.
\texttt{hosp.labevents} and \texttt{hosp.microbiologyevents} contain detailed raw laboratory and microbiology events, but these are not part of our extracted table.
We deliberately do not extract laboratory and microbiology event values from discharge summaries because all note timepoints are redacted and the note text is noisy and incomplete compared with the structured MIMIC event tables.
Instead, pertinent results provides a structured index over the clinician-selected results.
We segment the section into typed subsections each annotated with its line-range boundaries in the source text, the verbatim text they delimit, and per-type categorical tags such as imaging modality and body part.
We do not summarize or reinterpret the underlying values: the clinician's original wording, including any interpretation of radiology findings, is preserved verbatim.
However, a brief (3--5 word) topic label is added.
The section therefore adds a structured classification and indexing layer over clinically relevant results.

%% file: sections/appendix/cdm_benchmark.tex
\section{Experimental Details on CDM Benchmarks}
\label{app:cdm-experiments-details}

\begin{table*}[htb]
  \centering
  \small
  \caption{Diagnosis accuracy (\%) on the full-information and agentic CDM benchmarks, with 95\% Wilson confidence intervals in brackets.}
  \label{tab:cdm-confidence-intervals}
  \begin{tabular}{@{}l l c c@{}}
    \toprule
    \textbf{Model} & \textbf{Reasoning} & \textbf{Full information} & \textbf{Agentic} \\
    \midrule
    Gemini 3.5 Flash & minimal & 88.7 [87.4, 89.9] & \textbf{87.9} [87.2, 88.5] \\
    Gemini 3.1 Flash Lite & minimal & 87.4 [86.1, 88.7] & 81.9 [81.2, 82.7] \\
    Gemini 3.1 Flash Lite & high & 88.6 [87.3, 89.8] & 84.2 [83.5, 84.9] \\
    Gemma 4 31B & $\times$ & 86.7 [85.3, 88.0] & 83.7 [82.9, 84.4] \\
    Qwen3.6 35B-A3B & $\times$ & \textbf{90.4} [89.1, 91.5] & 83.6 [82.9, 84.3] \\
    Qwen3.6 35B-A3B & $\checkmark$ & 89.3 [88.0, 90.5] & 82.6 [81.8, 83.3] \\
    Qwen3.5 2B & $\times$ & 82.0 [80.4, 83.5] & 51.3 [50.2, 52.2] \\
    \bottomrule
  \end{tabular}
\end{table*}

\subsection{Cohort Construction}
\label{app:cdmv2-dataset-construction}

To construct the agentic CDM benchmark, we first identify admissions with a usable extracted primary discharge diagnosis and then apply a sequence of tabular filters to obtain a high-quality pool of diagnostic cases.
After filtering, 27{,}963 admissions remain as the candidate pool, from which we draw $T=10{,}000$ cases over the 19 classes (Table~\ref{tab:cdmv2-class-counts}).

\begin{table}[tb]
	\centering
	\scriptsize
	\setlength{\tabcolsep}{6pt}
	\caption{The 19 benchmark classes and the number of cases drawn.}
	\label{tab:cdmv2-class-counts}
	\begin{tabular}{lr@{\hskip 2em}lr}
		\toprule
		\textbf{Class} & \textbf{Cases} & \textbf{Class} & \textbf{Cases} \\
		\midrule
		Heart failure exacerbation & 731 & Acute cholecystitis & 646 \\
		Gastrointestinal bleeding & 731 & Appendicitis & 566 \\
		Cellulitis & 731 & Pancreatitis & 434 \\
		Small bowel obstruction & 731 & Cholangitis & 346 \\
		Pneumonia & 731 & Diverticulitis & 219 \\
		Urinary tract infection & 731 & Diabetic ketoacidosis & 172 \\
		COPD exacerbation & 730 & Aseptic meningitis & 159 \\
		Myocardial infarction & 730 & Pancreatic cancer & 120 \\
		Pulmonary embolism & 730 & Spontaneous bacterial peritonitis & 67 \\
		Colitis & 695 & & \\
		\bottomrule
	\end{tabular}
\end{table}

\subsection{Agent Harness}
\label{app:cdmv2-harness}

\paragraph{Interaction loop.}
Each case is evaluated separately with no memory retained between cases.
The agent is initialized with a system prompt that frames the diagnostic task and fixes the answer format, followed by the patient demographics and the triage chief complaint.
It then runs as a reason-and-act loop: it reasons, optionally issues one or more tool calls, reads the returned results, and repeats until it commits a final answer.
The candidate class names are supplied to the model, and the committed \enquote{Final diagnosis:} line is scored by case- and punctuation-insensitive substring matching.
We cap at ten tool-calling rounds.
On reaching the cap the agent is instructed to stop calling tools and output a diagnosis.

\paragraph{Tools and rendering.}
The agent interacts with each case through the five tools listed in Table~\ref{tab:cdmv2-tools}.
Each tool returns only the items the agent names, so the case is assembled incrementally.
Two tools additionally support a coarse-to-fine request pattern that keeps the initial view compact: the patient-record tool can return the list of available sections with their item counts (e.g.\ \texttt{conditions (7), allergies (2)}) before the agent asks for the content of a chosen section, and the laboratory tool can list the individual tests within a panel or body-fluid group before any are ordered.

Tool results are rendered as compact, uniform text:
laboratory values as \texttt{(Fluid) Test Label: Value | RR: [low -- high]} with the reference range appended when available;
physical-examination systems as \texttt{- System: Status - findings} lines preceded by a vital-signs line;
patient-record sections as headed bullet lists;
and imaging and clinical studies as the study title followed by the clinician-written report or result summary.
When a case never underwent a requested imaging or clinical study, the tool returns an explicit not-available message, and an already-returned study cannot be requested again.

\paragraph{Pertinent results.}
The \emph{pertinent results} section mixes study findings with physician decisions and diagnostic interpretations.
Fields containing diagnostic interpretation were excluded to prevent diagnosis leakage.
The remaining findings were divided between the \emph{imaging} and \emph{clinical studies} tools according to study type.

\begin{table}[htb]
	\centering
	\scriptsize
	\setlength{\tabcolsep}{4pt}
	\caption{The five tools exposed to the agent. Three are backed by \textsc{MIMIC-Note-Extract} (\textsc{MNE}); laboratory and imaging reuse existing MIMIC tables, with imaging additionally drawing on extracted pertinent-result imaging.}
	\label{tab:cdmv2-tools}
	\begin{tabular}{p{0.20\linewidth}p{0.15\linewidth}p{0.55\linewidth}}
		\toprule
		\textbf{Tool} & \textbf{Source} & \textbf{Requestable items} \\
		\midrule
		Physical examination & \textsc{MNE} & Six examination groups: vital signs, general, cardiopulmonary, abdominal, neurologic, and skin/extremities. \\
		Patient record & \textsc{MNE} & Eight sections: history of present illness, conditions, interventions, substance history, functional status, family history, allergies, and admission medications. \\
		Laboratory & MIMIC & Panels (CBC, basic metabolic, liver function, coagulation, urinalysis, blood gas, urine electrolytes), body-fluid analyses (CSF, ascitic, synovial, pleural), or individual tests from a catalog. \\
		Imaging & MIMIC + \textsc{MNE} & A region (abdomen, chest, head, neck, spine, extremity, venous) paired with a modality (CT, MRI, ultrasound, radiograph, nuclear medicine, fluoroscopy, ERCP). \\
		Clinical study & \textsc{MNE} & One study: echocardiogram, cardiac catheterization, electrocardiogram, stress test, upper endoscopy, or colonoscopy. \\
		\bottomrule
	\end{tabular}
\end{table}

\subsection{Models and Inference}
\label{app:cdm-models}

The models and inference code are shared by the agentic benchmark and the full-information setting.
We access the Gemini models~\citep{gemini-flash, gemini-flash-lite} through the Google Cloud Vertex AI API with zero data retention and serve the open-weight models locally with vLLM~\citep{kwon2023vllm}: Qwen3.6-35B-A3B~\citep{qwen36_35b_a3b}, a mixture of experts with 35B total and 3B active parameters, Gemma 4 31B~\citep{gemma42026}, and Qwen3.5-2B~\citep{qwen3_5}, all obtained from the Hugging Face Hub~\citep{wolf2020transformers}.
Qwen3.6-35B-A3B is served in FP8 (\texttt{Qwen/Qwen3.6-35B-A3B-FP8}), Gemma 4 31B in W4A16 (\texttt{google/gemma-4-31B-it-qat-w4a16-ct}), and Qwen3.5-2B unquantized.

\subsection{Agent Tool-Use Analysis}
\label{app:cdmv2-tool-usage}

We analyze how our fine-grained tools are used.
We find that tool-call behavior differs substantially across models.
Interestingly, increasing or enabling reasoning reduced the mean number of tool calls for both Gemini 3.1 Flash Lite and Qwen3.6 (Table~\ref{tab:cdmv2-tool-calls}).
In terms of tool call errors, Qwen3.5 2B frequently repeated requests and had the highest rates of malformed and excessive requests.

\begin{table}[htb]
  \centering
  \small
  \caption{Mean and median attempted tool calls per completed case, and percentage of calls with malformed or excessive requests.
  Excessive requests include repetitions and calls denied by request limits.}
  \label{tab:cdmv2-tool-calls}
  \begin{tabular}{@{}l l r r r r@{}}
    \toprule
    & & \multicolumn{2}{c}{\textbf{Calls per case}} & \multicolumn{2}{c}{\textbf{Requests (\%)}} \\
    \cmidrule(lr){3-4}\cmidrule(l){5-6}
    \textbf{Model} & \textbf{Reasoning} & \textbf{Mean} & \textbf{Median} & \textbf{Malformed} & \textbf{Excessive} \\
    \midrule
    Gemini 3.5 Flash & minimal & 6.95 & 7 & 2.01 & 1.85 \\
    Gemini 3.1 Flash Lite & minimal & 7.21 & 7 & 1.89 & 10.16 \\
    Gemini 3.1 Flash Lite & high & 3.73 & 4 & 0.92 & 0.39 \\
    Gemma 4 31B & $\times$ & 5.54 & 5 & 0.56 & 0.59 \\
    Qwen3.6 35B-A3B & $\times$ & 3.61 & 4 & 2.75 & 0.72 \\
    Qwen3.6 35B-A3B & $\checkmark$ & 2.95 & 3 & 5.30 & 0.16 \\
    Qwen3.5 2B & $\times$ & 9.09 & 11 & 8.10 & 30.64 \\
    \bottomrule
  \end{tabular}
\end{table}

Across all models, the agent's first tool call is almost always the patient record (88.7--100\% of cases) and the physical examination (92.2--100\%), and it queries each only once or twice per case.
Both are never requested in full: every physical-examination call names specific body systems, and each record read returns a median of three of the eight sections, typically after an initial call listing which sections exist.

The remaining tools are requested selectively, tracking what each disease requires (Figure~\ref{fig:cdmv2-tool-usage}, for the most accurate model, Gemini 3.5 Flash).
Imaging is ordered in 98.0\% of appendicitis, small-bowel obstruction, acute cholecystitis, cholangitis, diverticulitis, and pneumonia cases, where it is diagnostically decisive, and in 47\% of myocardial-infarction cases.
The clinical-study tool shows the sharpest targeting: 99.6\% of myocardial-infarction and 97.5\% of heart-failure cases, where an electrocardiogram is the decisive test, against 2--8\% of abdominal and skin pathologies.
This tool exists only because \textsc{MIMIC-Note-Extract} contains electrocardiograms and other studies from the pertinent-results extraction; the regex-derived benchmark could not expose them and cardiac pathologies were absent there.

\begin{figure}[tb]
	\centering
	\includegraphics[width=\linewidth]{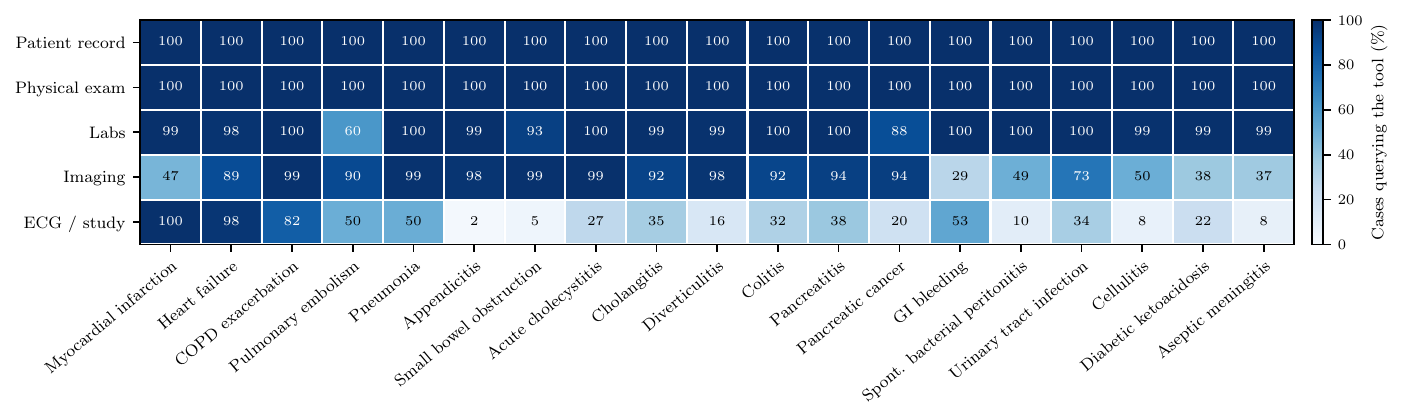}
	\caption{Fraction of cases in which Gemini 3.5 Flash queries each tool at least once, by pathology. The patient record and physical examination are queried almost universally, whereas imaging and the clinical-study (electrocardiogram, endoscopy) tool are requested selectively.}
	\label{fig:cdmv2-tool-usage}
\end{figure}

\paragraph{Disease-specific laboratory coverage.}
To assess whether agents obtained relevant diagnostic evidence, laboratory tests were grouped into disease-specific categories following \citet{hagerEvaluationMitigationLimitations2024}.
For pancreatitis, for example, these categories cover inflammation, pancreatic enzymes, and disease severity.
A case was counted when the agent obtained at least one laboratory result from each category.
Table~\ref{tab:cdmv2-lab-coverage} reports the percentage of cases meeting this criterion for four pathologies.

Coverage varied across models and diseases, with the largest differences observed for pancreatitis.
Interestingly, increasing reasoning in Gemini 3.1 Flash Lite reduced laboratory coverage while improving diagnosis accuracy, whereas enabling reasoning in Qwen3.6 reduced both coverage and accuracy (Table~\ref{tab:case-study-results}).
These results show that lower laboratory coverage does not necessarily imply lower diagnostic accuracy.

\begin{table}[htb]
  \centering
  \small
  \setlength{\tabcolsep}{4pt}
  \caption{Percentage of cases in which at least one laboratory result was obtained from every disease-specific category.
  Only cases completed by all model configurations and with results available from every category were included; counts are shown in the headings.}
  \label{tab:cdmv2-lab-coverage}
  \begin{tabular}{@{}l l r r r r@{}}
    \toprule
    \textbf{Model} & \textbf{Reasoning} & \shortstack{\textbf{Appendicitis}\\($n=345$)} & \shortstack{\textbf{Acute}\\\textbf{cholecystitis}\\($n=512$)} & \shortstack{\textbf{Diverticulitis}\\($n=216$)} & \shortstack{\textbf{Pancreatitis}\\($n=342$)} \\
    \midrule
    Gemini 3.5 Flash & minimal & 99.1 & 98.8 & 99.1 & 94.4 \\
    Gemini 3.1 Flash Lite & minimal & 94.5 & 97.7 & 94.9 & 78.7 \\
    Gemini 3.1 Flash Lite & high & 90.4 & 84.8 & 83.3 & 47.1 \\
    Gemma 4 31B & $\times$ & 89.6 & 87.7 & 81.9 & 57.9 \\
    Qwen3.6 35B-A3B & $\times$ & 88.4 & 80.3 & 76.4 & 52.3 \\
    Qwen3.6 35B-A3B & $\checkmark$ & 63.8 & 69.5 & 57.9 & 19.9 \\
    Qwen3.5 2B & $\times$ & 59.7 & 19.7 & 37.0 & 3.2 \\
    \bottomrule
  \end{tabular}
\end{table}